\documentclass{article}
\usepackage{iclr2026_conference}
\usepackage{graphicx}
\usepackage{amsmath,amssymb}
\usepackage{booktabs}
\usepackage{multirow}
\usepackage{verbatim}
\usepackage{hyperref}
\usepackage{tikz}
\usepackage{xcolor}

\newcommand{\imgbox}[6]{%
  \begin{tikzpicture}
    \node[anchor=south west, inner sep=0] (img) at (0,0)
      {\includegraphics[width=#1]{#2}};
    \begin{scope}[x={(img.south east)}, y={(img.north west)}]
      \draw[red, line width=1pt] (#3,#4) rectangle (#5,#6);
    \end{scope}
  \end{tikzpicture}%
}

\iclrfinalcopy
\usepackage{fancyhdr}

\title{Active World-Model with 4D-informed Retrieval for Exploration and Awareness}

\author{
\hspace*{-0.5em}Elaheh Vaezpour \\
\texttt{elaheh.vaezpour@kavai.com} \\
KavAI
\And
Amirhosein Javadi \\
\texttt{amjavadi@ucsd.edu} \\
UCSD
\And
Tara Javidi \\
\texttt{tjavidi@ucsd.edu} \\
KavAI \&
UCSD
}

\begin{document}
\maketitle

\vspace*{-.3in}
\begin{abstract}
Physical awareness, especially in a large and dynamic environment, is  shaped by sensing decisions that determine observability across space, time, and scale, while observations impact the quality of sensing decisions. This loopy information structure
makes physical awareness a fundamentally challenging decision problem with partial observations. While in the past decade we have witnessed the unprecedented success of reinforcement learning (RL) in problems with full observability, decision problems with partial observation, such as POMDPs, remain largely open: real-world explorations are excessively costly, while sim-to-real pipeline suffer from unobserved viewpoints. 

We introduce \textbf{AW4RE} (\textbf{A}ctive \textbf{W}orld-model with \textbf{4}D-informed \textbf{R}etrieval for \textbf{E}xploration), an awareness-centric 
generative world model 
that provides a sensor-native 
surrogate
environment for exploring sensing queries. 
Conditioned on a queried sensing action, AW4RE estimates the action-conditioned observation process. This is done by combining 4D-informed evidence retrieval, action-conditioned geometric support with temporal coherence, and conditional generative completion.
Experiments demonstrate that AW4RE produces more grounded and consistent predictions than geometry-aware generative baselines under extreme viewpoint shifts, temporal gaps, and sparse geometric support.
\end{abstract}

\vspace*{-.13in}
\section{Introduction}
\vspace*{-.13in}
Physical awareness, understanding what is happening in a real, dynamic environment across space and time, is a core requirement in robotics, autonomous driving platforms, and large-scale monitoring. Unlike classical perception, which infers semantics from a single viewpoint, physical awareness must integrate \emph{partial}, \emph{distributed}, and \emph{temporally asynchronous} observations into a coherent interpretation of the underlying world. In complex settings such as urban intersections, construction sites, or sports arenas, critical events often depend on geometry and interactions that unfold over time in multiple spatial scales, making awareness fundamentally a spatio-temporal inference problem.

A key property of physical awareness is that it is  \emph{observation-constraint} and \emph{action-conditioned}: what becomes observable depends on \emph{what} has already been observed and \emph{how} the environment is sensed. Camera placement, motion, orientation, and zoom determine which parts of the latent world state are revealed at each time, inducing partial observability by design. This naturally leads to a decision-theoretic formulation: an agent selects sensing actions to reduce uncertainty about a latent spatio-temporal state while respecting sensing costs. In this paper, we formalize this setting as a finite-horizon spatio-temporal Partially Observable Markov Decision Process (POMDP), where actions parameterize an observation process and the objective is to optimize camera policies under partial observability.

Reinforcement learning has succeeded to solve decision problems in fully observable settings, but policy learning under partial observability, as in POMDPs, remains open. In real-world settings, using the physical environment directly for reinforcement learning is infeasible: Camera actions are costly, slow, and often irreversible, and partial observability implies that poor sensing decisions may permanently hide the evidence needed to evaluate alternative policies. As a result, relying on real-world interaction alone does not provide a scalable learning signal. Sim-to-real pipelines (\cite{abou2025real, wu20244d}) offer an alternative, but they inherit blind spots introduced by unobserved viewpoints and incomplete sensing coverage. Scalable camera-policy optimization therefore requires a sensor-native \emph{surrogate environment} that can support counterfactual reasoning—answering questions of the form: \emph{What would have been observed under a different sensing action?}

Recent generative world models and geometry-aware video diffusion methods (\cite{agarwal2025cosmos, bruce2024genie, assran2025vjepa}) provide promising ingredients, but they are not designed to function as reliable environments for physical awareness. These models often prioritize visual plausibility and short-horizon continuity, and they degrade when queried under substantial viewpoint shifts, scale changes, or temporal gaps. The central challenges are fidelity under sparse spatio-temporal evidence and world-consistency under a large corpus of evidence (volumetric context). It is clear that in large-scale and complex environments, the model must distinguish which parts of a queried view are already constrained by available observations and which remain uncertain, and it must avoid hallucinating content that is supported by evidence.

In contrast, we utilize our reinforcement learning framework and our proposed factorization to introduce AW4RE (Active World model for 4D Retrieval and Exploration), an awareness-centric world model that combines action-conditioned 4D-informed evidence retrieval, local geometric support with temporal coherence, and conditional generative completion. Rather than maintaining a global 4D reconstruction, AW4RE operates locally at the level of individual query frames. For each queried sensing action, it retrieves a compact set of spatio-temporally relevant observations, constructs a local geometric proxy for evidence-supported regions, and projects this proxy into the query camera to produce a partially filled frame. A conditional video diffusion model then completes the remaining regions in a temporally coherent manner, guided by the explicit separation between evidence-supported and unsupported content. 

While  AW4RE is a general framework for physical awareness, we demonstrate the power of the framework by implementing it in the context of RGB observation and 4D novel-view rendering.
More specifically, we evaluate AW4RE as a surrogate environment for counterfactual sensing on the Waymo Open Dataset (\cite{sun2020scalability}) and compare it against GEN3C (\cite{ren2025gen3c}), the state of the art geometry-aware generative approach. The evaluation uses queries that stress multi-view reasoning, multi-scale coherence, temporal extrapolation beyond the observed window, and sparse geometric support. Across these settings, the proposed 4D-informed evidence selection provides stronger constraints for generation than time-local conditioning, leading to more consistent geometry and reduced hallucination under partial observability. Our experiments show that AW4RE provides a physical awareness-centric world 
models that explicitly reason about how observations depend on sensing actions. Our model supports counterfactual evaluation grounded in prior evidence: given a history of partial observations, the model should predict what would be observed under a different camera configuration, even when that viewpoint, scale, or time was never directly seen.

\vspace*{-.13in}
\section{Related Approaches}
\label{sec:related_envs}
\vspace*{-.13in}
Several classes of approaches attempt to address the need for an environment for exploration and learning sensing policies.

\vspace*{-.13in}
\paragraph{4D reconstruction methods.}
Techniques such as 4D reconstruction and dynamic scene modeling reconstruct environments from dense multi-view video~(\cite{wu20244d,park2021hypernerf}). These representations allow replay of observed events, but require comprehensive capture in advance and cannot reason about unobserved viewpoints or missing details. Consequently, reasoning beyond the captured observations requires an explicit model of geometry and dynamics, motivating model-driven approaches such as digital twins.

\vspace*{-.13in}
\paragraph{Digital twins.}
To overcome the limitations of data-driven reconstruction, digital twins simulate environments using predefined geometry, physics, and system rules~(\cite{iranshahi2025digital,abou2025real}). Although they enable controlled experimentation beyond recorded data, they are expensive to build, require significant manual effort, and are difficult to adapt to changing or unseen environments. Moreover, they do not learn from data and therefore cannot infer unobserved structure.

Despite their differences, both classes of approaches are fundamentally limited to replaying or simulating what has been explicitly observed or manually modeled. As a result, they do not provide a scalable environment for learning camera actions or evaluating sensing policies under partial observability. These limitations motivate approaches that can learn environment structure and dynamics directly from data rather than relying on manually specified models.

\vspace*{-.13in}
\paragraph{World models.}
Recent world models learn environment structure and dynamics directly from data, enabling future prediction and generation. 
Together, these models represent a significant step toward learned environments and imagination-based reasoning.
V-JEPA 2-AC~(\cite{assran2025vjepa}) is an action-conditioned world model that takes as input video patches paired with numerical 7D vectors representing the robot’s physical state and proposed actions. Its primary application is robotic planning, where the model imagines future states in a latent space to evaluate action sequences that achieve a specified visual goal.
Google Genie~(\cite{bruce2024genie}) is a world model that generates interactive environments from a single image and a text prompt. The model allows users to navigate and interact with the generated world in real time, simulating complex physical transitions without relying on an explicit physics engine.
NVIDIA Cosmos~(\cite{agarwal2025cosmos}) is a suite of physics-aware diffusion models that synthesize high-fidelity video from text, images, and action-conditioned inputs. By representing actions as latent variables within the diffusion process, the model can simulate complex physical interactions and evaluate robotic policies in an environment represented as video. These world models are powerful within their intended application domains. However, they do not consider camera action with physical limitations dictated by camera intrinsics and extrinsics. Furthermore, they operate under limited temporal context, restricted viewpoint variation, or fixed observation scales. Hence, they often struggle when reasoning about viewpoints, resolutions, or time horizons that differ substantially from those observed. 

\vspace*{-.13in}
\paragraph{Geometry-aware conditional video diffusion models.}
Geometry-aware conditional video diffusion models extend video generation by conditioning explicitly on camera parameters, depth, or 3D structure, enabling controlled viewpoint changes and multi-view synthesis (\cite{kuang2024collaborative, hou2024training, ren2025gen3c}). By grounding generation in geometric cues, these models improve spatial coherence across views and time, making them better suited for reasoning about camera motion than purely image- or action-conditioned world models. However, existing geometry-aware diffusion models are still optimized primarily for visual fidelity and short-horizon consistency. More specifically, their performance typically relies on strong overlap between observed and generated content, including overlapping viewpoints, comparable spatial scales, and aligned temporal context. When asked to generate views that differ substantially in viewpoint, resolution, or timing from the available observations, these models often drift or hallucinate, limiting their ability to support awareness-driven reasoning or counterfactual evaluation of camera actions under partial observability.

Motivated by these limitations, we propose in the next sections a learned environment that enables awareness-driven reasoning and the learning of camera actions under partial observability.

\vspace*{-.163in}
\section{Physical Awareness as a Spatio-Temporal POMDP}
\label{sec:pomdp}
\vspace*{-.13in}
We model physical awareness in a real, dynamic environment as a POMDP with a finite time horizon. The environment exists over a discrete time interval $t \in \{1,\dots,T\}$ and corresponds to a continuous 4D spatio-temporal physical world.

\vspace*{-.13in}
\paragraph{POMDP definition.}
The process is defined by the tuple 
$
\mathcal{M} = \langle \mathcal{S}, \mathcal{A}, \mathcal{O}, T, Z, R, \gamma \rangle
$
where the agent optimizes a sensing (camera) policy under partial observability.

\vspace*{-.13in}
\paragraph{Latent state.}
At each time $t$, the latent state $s = (s_1, s_2, \dots, s_T) \in \mathcal{S}$ where each $s_t$ represents the true physical state of the environment at time $t \in \{1,\dots,T\}$. The state $s$ therefore corresponds to the full spatio-temporal trajectory of the environment over the time horizon. 

\vspace*{-.13in}
\paragraph{Action space.}
The agent interacts with the environment by selecting sensing actions. At each iteration $k$, the agent chooses a sequence of sensing actions $
a^{(k)} = \{a^{(k)}_t \mid t \in \{1,\dots,T\}\},
$ 
where each action $a^{(k)}_t \in \mathcal{A}$ specifies a sensing configuration at time $t$. For cameras as sensors, it can include position, orientation, zoom level, and other imaging parameters. Actions do not alter the physical environment, but determine which aspects of the latent state become observable.

\vspace*{-.13in}
\paragraph{State dynamics.}
The dynamic component of the state evolves according to unknown physics-constrained transition dynamics: $s_{t+1} \sim T(s_{t+1} \mid s_t)$.
These dynamics capture motion, interactions, and other physical changes and are independent of sensing actions.

\vspace*{-.13in}
\paragraph{Observation model.}
Observations are generated according to an action-dependent observation function. For a given sensing action indexed by $k$, the observation is drawn as 
\begin{equation}
o^{(k)} \sim Z(\cdot \mid s, a^{(k)}),
\end{equation}
where $o^{(k)} \in \mathcal{O}$ denotes the sensor output at iteration $k$. Given the conditional independence across time conditioned on the latent state trajectory and sensing actions, we have 
\begin{equation}
p(o^{(k)} \mid s, a^{(k)}) = \prod_{t=1}^{T} Z\!\left(o^{(k)}_t \mid s_t, a^{(k)}_t\right).
\end{equation}

\vspace*{-.13in}
The key property is the factorization under action-dependence of the observations. 

\vspace*{-.133in}
\paragraph{Observation history.}
After $K$ iterations, $\mathcal{D}^{(k)} = \{(o^{(j)}, a^{(j)})\}_{j=1}^{k-1}$ denotes the history.

\vspace*{-.13in}
\paragraph{Reward.}
The reward encourages reducing uncertainty about the environment while accounting for sensing costs:

\begin{equation}
R^{(k)} = R^{(k)}_{task} + \mathcal{I}(s; o^{(k)} \mid a^{(k)}, \mathcal{D}^{(k)}) - \lambda \, \text{Cost}(a^{(k)}),
\end{equation}
where $\mathcal{I}$ denotes the expected information gain, $\mathrm{Cost} (a^{(k)})$ captures constraints, and task-specific reward $R^{(k)}_{task}$ (e.g., safety or anomaly detection) can be incorporated when available.

\vspace*{-.13in}
\paragraph{Objective.}
The camera policy which maps observations to action, $\pi(. \mid D^{(k)})$, is optimized to maximize expected cumulative reward: $ \max_{\pi} \mathbb{E}_{\pi}\left[\sum_{k=0}^{K} \gamma^k R^{(k)}\right]$.

\section{AW4RE}
\label{sec:method}

\vspace*{-.13in}
Direct access to the true state transition dynamics and the action-conditioned observation model of the environment is unavailable. Instead, we observe only a sparse set of spatio-temporal views generated by previously executed sensing actions. These observations provide partial and asynchronous samples of the underlying observation process, making direct policy learning and counterfactual evaluation challenging \cite{IAUP}.
To address this issue, we therefore introduce AW4RE, an \emph{awareness-centric world model} that estimates the action-conditioned observation process indirectly, using accumulated evidence from prior sensing actions rather than the latent state itself. Given a candidate sensing action $a^{(k)}$, our goal is to estimate the observation distribution $Z(o^{(k)} \mid s, a^{(k)})$. 

As formalized in Eq.~\eqref{eq:world_model}, we decompose this estimation into three tightly coupled components. First, a \emph{4D-informed state estimator module} $\hat{Z}_1$ aggregates prior observations and actions to produce an internal summary/estimate $\hat{s}$ of the latent spatio-temporal state. Second, an \emph{evidence-backed action-conditioned observation module} $\hat{Z}_2$ uses this inferred state conditioned on the queried sensing action to construct an action-conditioned evidence-adjusted partially supported observation $\tilde{o}^{(k)}$. Finally, a \emph{conditional generative observation module} $\hat{Z}_1$ completes the unsupported regions to produce the final estimated observation $\hat{o}^{(k)}$.

\vspace*{-.13in}
\begin{equation} \label{eq:world_model}
Z(o^{(k)} \mid s, a^{(k)}) \;\approx\; 
\hat{Z}_3 (\hat{o}^{(k)} \mid \tilde{o}^{(k)} )
\hat{Z}_2 (\tilde{o}^{(k)} \mid \hat{s}, a^{(k)})
\hat{Z}_1 (\hat{s} \mid \hat{o}^{(1:k-1)}, a^{(1:k)}),
\end{equation}
which predicts the observation that would result from executing sensing action $a^{(k)}$, conditioned on all previously acquired evidence.


As additional sensing actions are evaluated, the evidence set $\mathcal{D}^{(k)}$ grows, progressively constraining the learned approximation and improving the fidelity of predicted observations. When evidence from a different environment is provided, the conditioning set changes accordingly, allowing the same world model to support counterfactual reasoning across environments without manual reconstruction or retraining. In this way, AW4RE serves as a queryable surrogate for the action-conditioned observations of the underlying POMDP.

\vspace*{-.13in}
\paragraph{Visual Case Study.} An important special case of our framework arises wehn sensors provide visual (RGB) information. In that setting, each action $a^{(k)} = \{a^{(k)}_t\}_{t=1}^{T}$ corresponds to the camera configurations at iteration $k$ where each $a^{(k)}_t$ specifies a camera configuration at time $t$, consisting of the intrinsics and extrinsics of the camera. 
In the camera-based setting, each observation $o^{(k)}_t \in \mathcal{O}$ is an RGB image captured at time $t$ under camera action $a^{(k)}_t$. 
A full observation at iteration $k$ is the resulting video $o^{(k)} = \{o^{(k)}_t\}_{t=1}^{T}$. 
At iteration $k$, the available evidence consists of previously collected RGB videos and their associated camera configurations, $\mathcal{D}^{(k)} = \{(o^{(j)}, a^{(j)})\}_{j=1}^{k-1}$. Given a camera action $a^{(k)} = \{a^{(k)}_t\}_{t=1}^{T}$, our goal is to generate the corresponding RGB video $o^{(k)} = \{o^{(k)}_t\}_{t=1}^{T}.$

\vspace*{-.13in}
\paragraph{4D-informed State Estimator Module $\hat{Z}_1(\hat{s} \mid \hat{o}^{(1:k-1)}, a^{(1:k)})$.}
The state estimator module aggregates accumulated spatio-temporal evidence to form an internal summary/estimate, also known as sufficient statistics, of the latent world state. One natural approach is to construct a global 4D reconstruction from the entire observation history $\mathcal{D}^{(k)}$. However, as the number of sensing iterations grows, the observation history $\mathcal{D}^{(k)}$ expands unboundedly in both time and viewpoint coverage. 

Rather than constructing a global 4D reconstruction from all frames in $\mathcal{D}^{(k)}$, we instead operate at the level of individual queried camera actions. For each queried time step $t$, the action $a^{(k)}_t$ specifies a camera configuration, and our goal is to determine which past RGB frames contribute support to the corresponding view. Concretely, for each queried frame $o^{(k)}_t$, we select a subset of past frames from $\mathcal{D}^{(k)}$ whose camera viewpoints and timestamps are expected to provide 3D constraints for the region visible from $a^{(k)}_t$. Using only this selected subset, we build a sequence of $T$ local 3D point cloud that captures the portion of the scene supported by past evidence for the query view.

Let the evidence corpus at iteration $k$ be
\begin{equation}
\mathcal{D}^{(k)}=\{(o^{(j)}_{i}, a^{(j)}_{i}) \;:\; j\in\{1,\dots,k-1\},\; i\in\{1,\dots,T\}\},
\end{equation}
where $o^{(j)}_{i}$ is an RGB frame and $a^{(j)}_{i}$ is its associated camera intrinsics and extrinsics.
Given a queried camera action $a^{(k)}_t$, we seek a compact set of evidence frame indices
\begin{equation}
\mathcal{I}_t \subset \{1,\dots,k-1\}\times\{1,\dots,T\}, 
\qquad |\mathcal{I}_t|\le M,
\end{equation}
that specifies which frames in the corpus will be used to support the reconstruction of the query frame at $a^{(k)}_t$.
We refer to the corresponding retrieved context as
\begin{equation}
\mathcal{C}_t(\mathcal{I}_t)
=
\{(o^{(j)}_{i}, a^{(j)}_{i}) \;:\; (j,i)\in\mathcal{I}_t\}.
\end{equation}

\vspace*{-.13in}
We choose $\mathcal{I}_t$ to maximize a relevance score measuring how much the selected frames are expected to contribute to the query view:
\begin{equation}
\mathcal{I}_t^\star
=
\arg\max_{\mathcal{I}_t:\,|\mathcal{I}_t|\le M}
\;\;\sum_{(j,i)\in\mathcal{I}_t}
\Omega\!\big((o^{(j)}_{i}, a^{(j)}_{i}),\, a^{(k)}_t\big),
\qquad t=1,\dots,T,
\label{eq:index_select}
\end{equation}
where $\Omega$ is an action-conditioned contribution metric.

Given a queried camera action $a^{(k)}_t$ and retrieved evidence  $\mathcal{C}_t(\mathcal{I}_t^\star)$, we form local 3D support for the query view by (i) using \emph{all} pixels from evidence frames captured at the \emph{same} time $t$ (to preserve dynamic content), and (ii) using only the \emph{static} components from evidence frames captured at \emph{different} times (to avoid injecting inconsistent dynamics). Concretely, 
\begin{equation}
\mathcal{I}^{\star^\mathrm{on}}_t = \{(j,i)\in\mathcal{I}_t : i=t\}, 
\qquad
\mathcal{I}^{\star^\mathrm{off}}_t = \mathcal{I}_t \setminus \mathcal{I}^{\star^\mathrm{on}}_t,
\end{equation}
partitions the indices and defines a time-aware operator
\vspace*{-.05in}
\begin{equation}
\phi^{(k)}_t
=
\Phi \!\left(
a^{(k)}_t;\, \mathcal{C}_t(\mathcal{I}_t)
\right),
\label{eq:mix_project}
\end{equation}
where $\Phi$ constructs a local 3D proxy from the union of
\begin{equation}
\{(o^{(j)}_{t}, a^{(j)}_{t})\}_{(j,t)\in\mathcal{I}^{\star^\mathrm{on}}_t}
\;\;\cup\;\;
\{(o^{(j)}_{i,\mathrm{stat}}, a^{(j)}_{i})\}_{(j,i)\in\mathcal{I}^{\star^\mathrm{off}}_t}.
\end{equation}
Here, $o^{(j)}_{i,\mathrm{stat}}=(1-m^{(j)}_{i})\odot o^{(j)}_{i}$ denotes the static-masked version of a cross-time frame, obtained by removing dynamic regions via a mask $m^{(j)}_{i}$. This design allows cross-time evidence to contribute stable geometric scaffolding, while same-time evidence contributes both static and dynamic content when available. The output of this module therefore constitutes an \emph{action-conditioned, 4D-informed estimate of the environment state}, represented as local 3D point clouds that summarize the portions of the environment supported by prior evidence.

\vspace*{-.13in}
\paragraph{Evidence-backed Observation Module $\hat{Z}_2(\tilde{o}^{(k)} \mid \hat{s}, a^{(k)})$.}
The local 3D proxy $\phi^{(k)}_t$ represents the subset of the scene that is supported by prior evidence for $a^{(k)}_t$. To obtain an evidence-backed partial observation in the queried camera configuration, we define a single \emph{action-aware operator} that jointly performs geometric projection and support densification:
\begin{equation}
\tilde{o}^{(k)}_t
=
\Omega\!\left(\phi^{(k)}_t,\; a^{(k)}_t\right),
\label{eq:action_aware_decode}\vspace*{-.053in}
\end{equation}
where $\Omega$ maps the local 3D proxy $\phi^{(k)}_t$ into the image plane specified by the camera intrinsics and extrinsics $a^{(k)}_t$, while adapting to the queried camera action. When 4D-informed support is sufficient, $\Omega$ reduces to standard projection, populating pixels that are directly constrained by evidence. When support is sparse (e.g., under zoom-in queries or limited viewpoint overlap), $\Omega$ incorporates an action-conditioned densification step (e.g., learned upsampling or scale-aware smoothing) to produce a dense, low-frequency guidance signal.
The resulting $\tilde{o}^{(k)}_t$ is a partially decoded RGB frame in which evidence-supported regions are populated in a 4D-informed manner, while unsupported regions remain explicitly unconstrained. This decoding procedure is applied independently for each time step $t = 1,\dots,T$, yielding a partially observed video $\tilde{o}^{(k)} = \{\tilde{o}^{(k)}_t\}_{t=1}^{T}$.
By action-conditioned projection, the model produces stable and informative guidance for subsequent generative completion. 

\vspace*{-.13in}
\paragraph{Conditional Generative Observation Module. $\hat{Z}_3 (\hat{o}^{(k)} \mid \tilde{o}^{(k)} )$.}

We use a conditional video diffusion model to complete the missing regions of the video in a visually realistic and temporally coherent manner. By conditioning on per-frame geometric support, this approach grounds generation in the most relevant past observations.
In other words, the partially supported frames
\(
\tilde{o}^{(k)} = \{\tilde{o}^{(k)}_t\}_{t=1}^{T}
\)
are passed to a conditional video diffusion model. This module completes the unsupported regions while preserving temporal coherence, producing the final estimated video
\(
\hat{o}^{(k)} = \{\hat{o}^{(k)}_t\}_{t=1}^{T}.
\)

\vspace*{-.13in}
\section{Experimental Results}
\label{sec:results}
\vspace*{-.13in}
We evaluate AW4RE through a series of quantitative and qualitative comparisons. 

\vspace*{-.13in}
\paragraph{Comparison setup.}
We compare our world model against GEN3C (\cite{ren2025gen3c}), which represents the state-of-the-art geometry-aware generative baseline, using sequences from the Waymo Open Dataset (\cite{sun2020scalability}). Both models are provided with identical partial observation histories and are queried with the same counterfactual sensing actions. These actions include large viewpoint shifts, changes in camera scale (near vs.\ far), and temporal extrapolation beyond the observed window  enabling a controlled evaluation of reasoning under sparse and incomplete spatio-temporal evidence. For the conditional video diffusion model, we use the fine-tuned version of Cosmos1.0Diffusion7BVideo2World\footnote{\url{https://github.com/nv-tlabs/GEN3C/tree/main/cosmos_predict1}}.

\vspace*{-.13in}
\paragraph{Performance Metrics.} We evaluate both spatial fidelity and temporal stability using complementary image- and video-level metrics. When ground-truth (GT) frames are available, we report full-frame PSNR (\cite{wang2004image}), SSIM (\cite{wang2004image}), and LPIPS (\cite{zhang2018unreasonable}) to measure overall observation estimation quality. In addition, we evaluate evidence quality, defined over evidence-backed regions of the image, i.e., pixels whose appearance is constrained by previous observations. When these regions form a contiguous spatial support, 
we additionally report SSIM and LPIPS over the region. To assess temporal consistency, we report temporal LPIPS (T-LPIPS) (\cite{zhang2018unreasonable}), computed between consecutive frames, both over the full frame and restricted to evidence-backed regions. Lower T-LPIPS indicates improved temporal stability. When GT is unavailable, evaluation is restricted to evidence-backed regions and temporal metrics, as discussed below.

\vspace*{-.12in}
\paragraph{Partial Temporal Evidence.}
We evaluate temporal reasoning under partial observability using a fixed camera viewpoint with missing observations. 
When only early frames are observed and later frames are masked as shown in Fig.~\ref{fig:play_forward}, where GT frames are available, and quantitative results are presented in Table \ref{tab:gt_temporal}. We see that AW4RE consistently outperforms GEN3C across all full-frame quality metrics, achieving substantially higher PSNR and SSIM and significantly lower LPIPS. The gains are even more pronounced over evidence-backed regions, where AW4RE leverages 4D-informed retrieval to tightly constrain geometry and appearance, resulting in higher spatial fidelity compared to time-local conditioning. Temporal stability results further highlight this advantage: AW4RE achieves lower full-frame and evidence-region T-LPIPS than GEN3C. This indicates that AW4RE not only improves per-frame quality but also maintains coherent structure and appearance over time, even under partial observability.


\begin{table}[htbp]
\centering
\caption{Partial Temporal Evidence Evaluation}
\label{tab:gt_temporal}
\setlength{\tabcolsep}{3pt}
\resizebox{\textwidth}{!}{%
\begin{tabular}{@{}llcccccccc@{}}
\toprule
 & & \multicolumn{3}{c}{\textbf{Full-frame Metrics}} & \phantom{a} & \multicolumn{3}{c}{\textbf{Evidence Metrics}} \\
\cmidrule{3-5} \cmidrule{7-9}
\textbf{Camera Action} & \textbf{Model} & \textbf{PSNR} $\uparrow$ & \textbf{SSIM} $\uparrow$ & \textbf{LPIPS} $\downarrow$ & & \textbf{PSNR} $\uparrow$ & \textbf{SSIM} $\uparrow$ & \textbf{LPIPS} $\downarrow$ \\ \midrule
\textbf{Temporal Query (Next Time Steps)}  & AW4RE & \textbf{28.532} & \textbf{0.905} & \textbf{0.100} & & \textbf{31.864} & \textbf{0.927} & \textbf{0.057} \\
                          & GEN3C & 15.269          & 0.467          & 0.602          & & 13.520          & 0.411          & 0.626          \\ \midrule
\textbf{Temporal Query (Previous Time Steps)} & AW4RE & \textbf{32.265} & \textbf{0.927} & \textbf{0.058} & & \textbf{32.768} & \textbf{0.919} & \textbf{0.049} \\
                          & GEN3C & 26.677          & 0.780          & 0.211          & & 26.111          & 0.717          & 0.233          \\ \midrule
\midrule
 & & \multicolumn{7}{c}{\textbf{Temporal Consistency}} \\ \cmidrule{3-9}
 & & \multicolumn{3}{c}{\textbf{Full T-LPIPS} $\downarrow$} & \phantom{a} & \multicolumn{3}{c}{\textbf{Evidence T-LPIPS} $\downarrow$} \\ 
\cmidrule{3-5} \cmidrule{7-9}
\textbf{Temporal Query (Next Time Steps)}  & AW4RE & \multicolumn{3}{c}{\textbf{0.0118}} & & \multicolumn{3}{c}{\textbf{0.0151}} \\
                          & GEN3C & \multicolumn{3}{c}{0.1635}          & & \multicolumn{3}{c}{0.1758}          
                          \\ \midrule
\textbf{Temporal Query (Previous Time Steps)} & AW4RE & \multicolumn{3}{c}{\textbf{0.0066}} & & \multicolumn{3}{c}{\textbf{0.0065}} \\
                          & GEN3C & \multicolumn{3}{c}{0.0388}          & & \multicolumn{3}{c}{0.0276}           
                        \\  \bottomrule
\end{tabular}%
}
\end{table}

\begin{figure*}[t]
\centering
\setlength{\tabcolsep}{4pt}

{\footnotesize
\begin{tabular}{c|ccccc}
 & $o^{(1)}_1$ & $o^{(1)}_{20}$ & $o^{(1)}_{40}$ & $o^{(1)}_{50}$ & $o^{(1)}_{70}$ \\
\rotatebox{90}{Evidence}
& \includegraphics[width=0.17\linewidth]{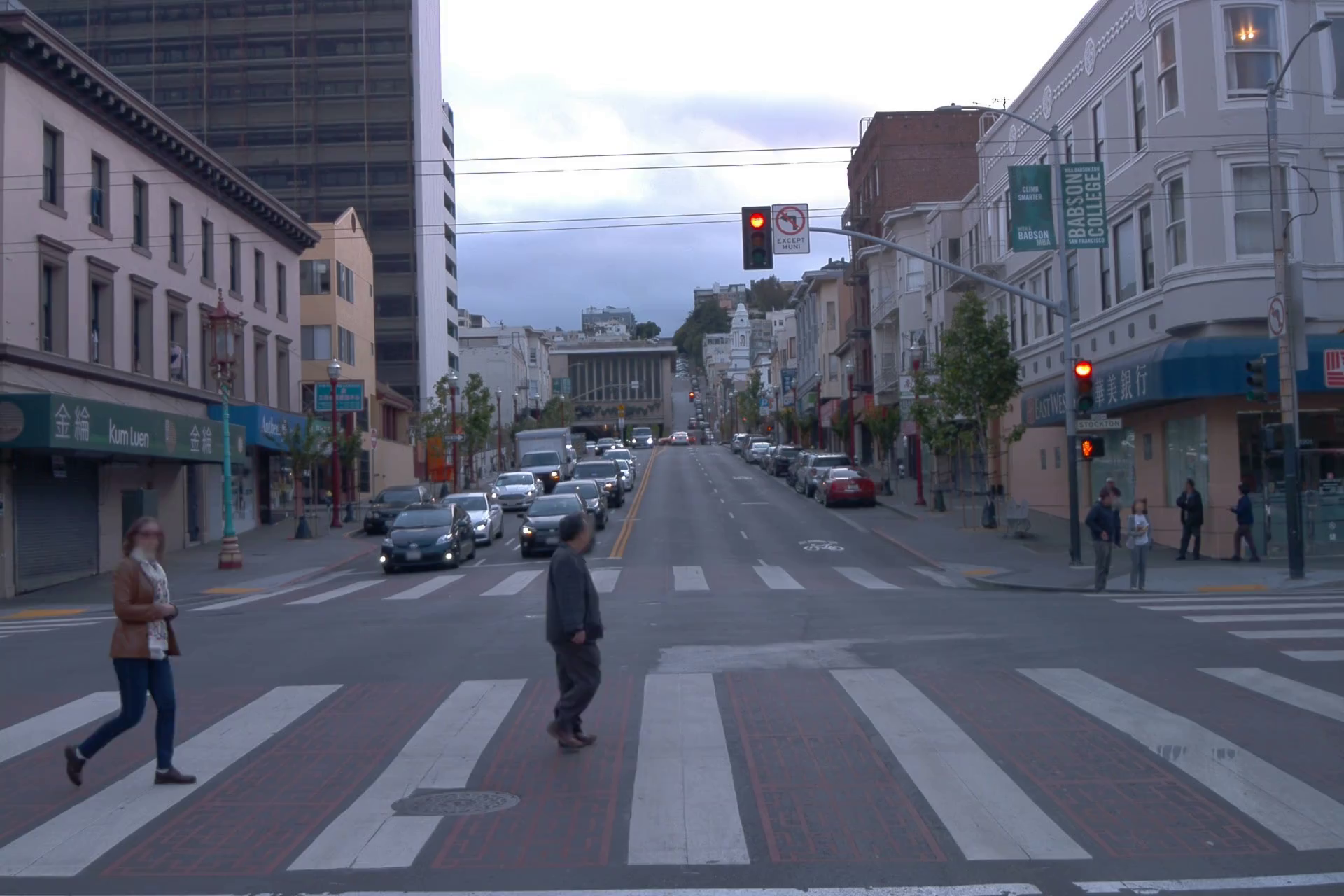}
& \includegraphics[width=0.17\linewidth]{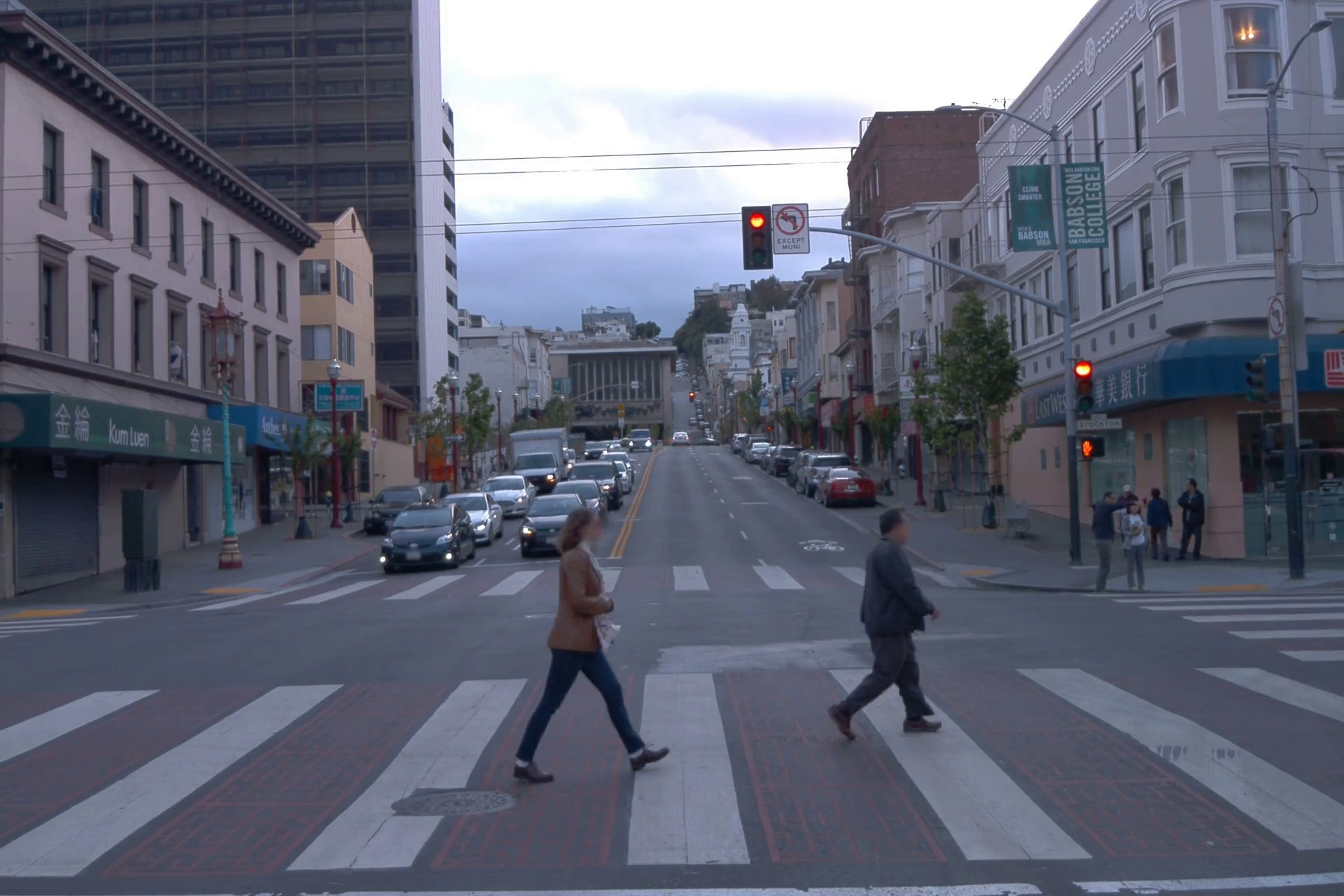}
& \includegraphics[width=0.17\linewidth]{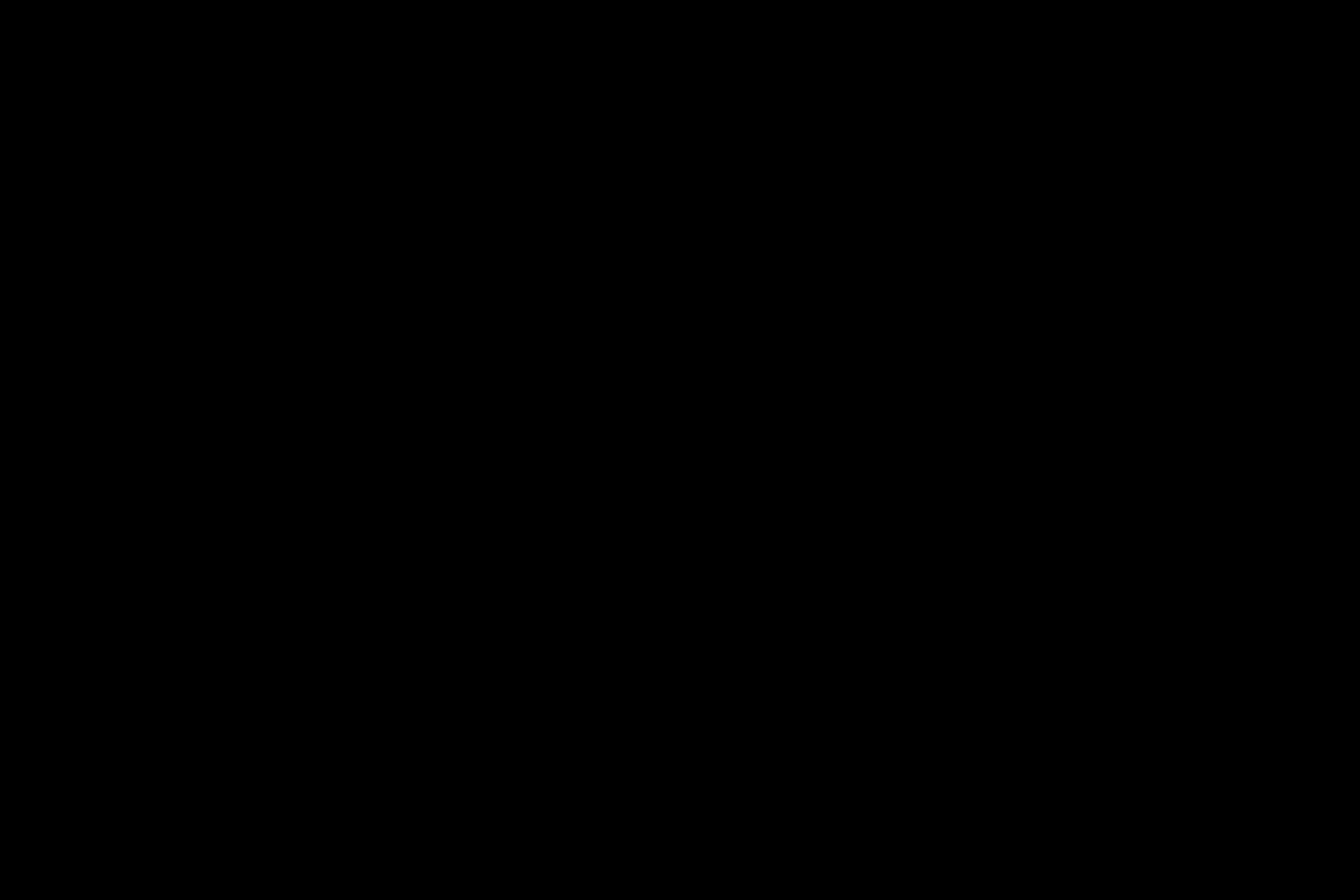}
& \includegraphics[width=0.17\linewidth]{figs/test_5/input_video/black.png}
& \includegraphics[width=0.17\linewidth]{figs/test_5/input_video/black.png}
\\
& \multicolumn{5}{c}{} \\
 & \multicolumn{5}{c}{$\mathcal{D}^{(1)} = \{(\{o^{(1)}_t\}_{t=1}^{121}, \{a^{(1)}_t\}_{t=1}^{121})\}$, \   
$ a^{(2)}_t = a^{(1)}_t \  \forall t \in \{1,\dots,20\}, 
\quad
a^{(2)}_t = a^{(1)}_{20} \  \forall t \in \{21,\dots,121\}.
$} \\
 & \multicolumn{5}{c}{} \\
 & $o^{(2)}_1$ & $o^{(2)}_{20}$ & $o^{(2)}_{40}$ & $o^{(2)}_{50}$ & $o^{(2)}_{70}$ \\

\rotatebox{90}{GEN3C}
& \includegraphics[width=0.17\linewidth]{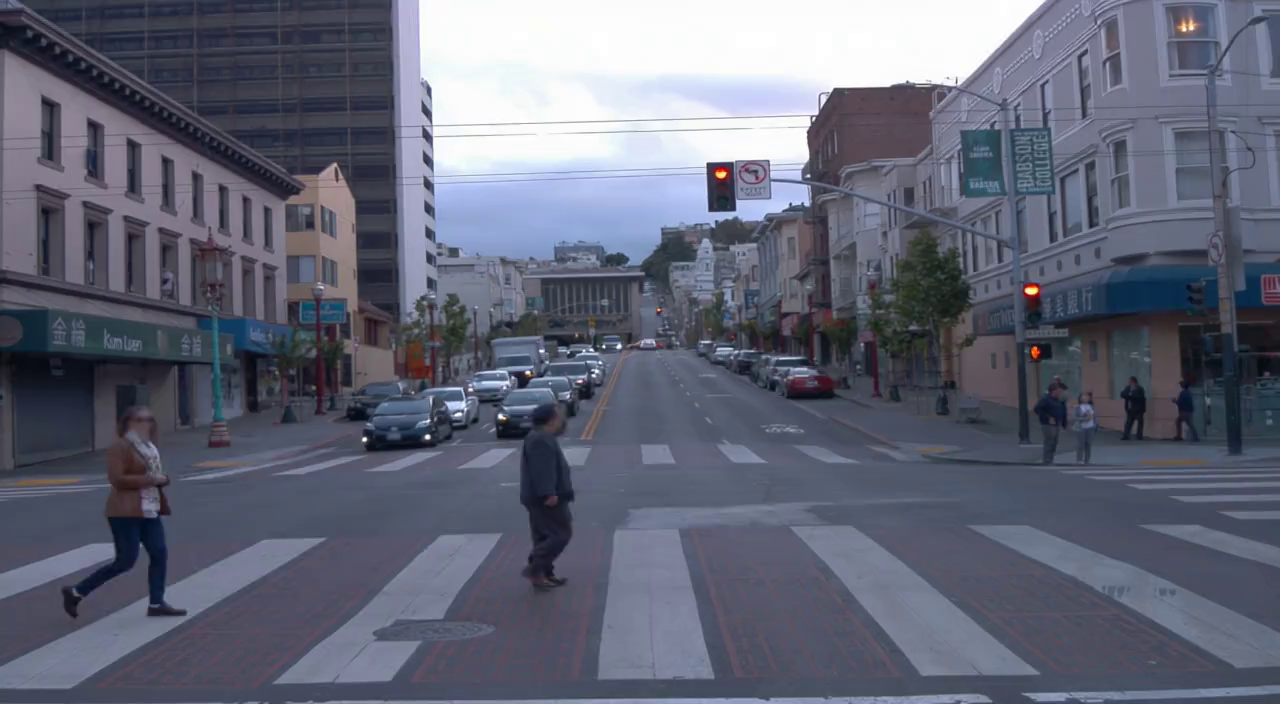}
& \imgbox{0.17\linewidth}{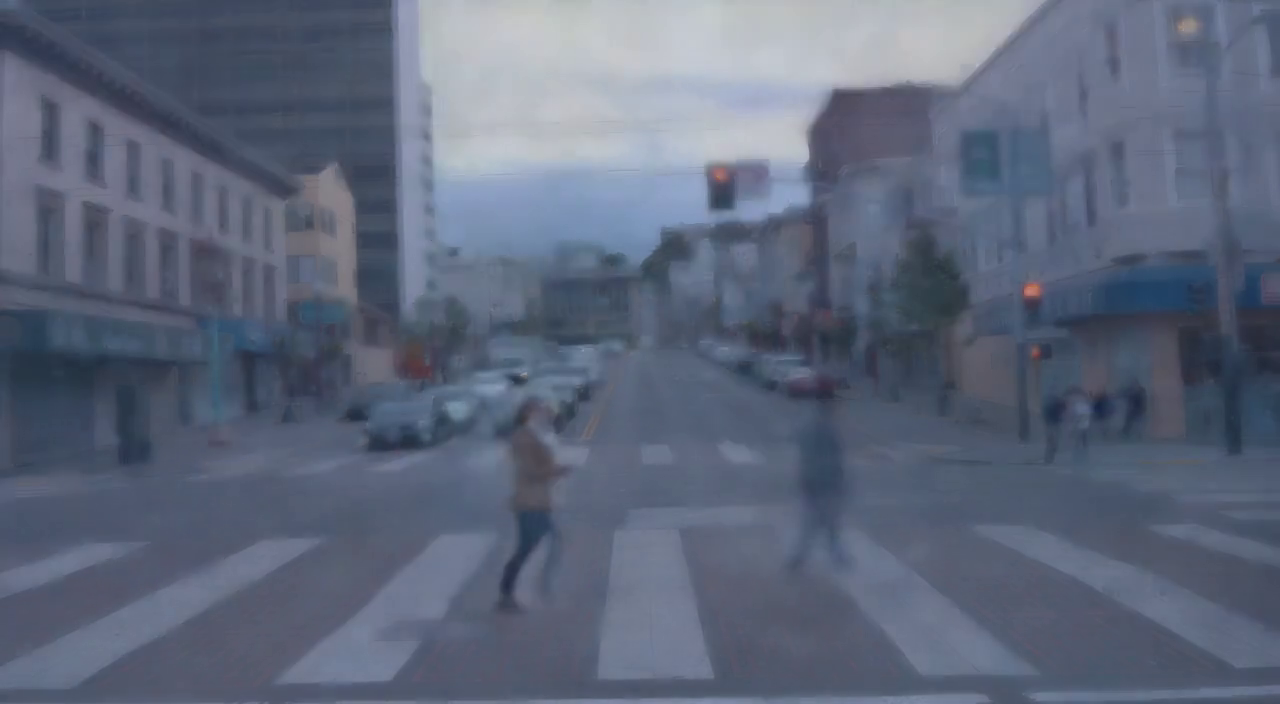}{0.3}{0.10}{0.75}{0.65}
& \imgbox{0.17\linewidth}{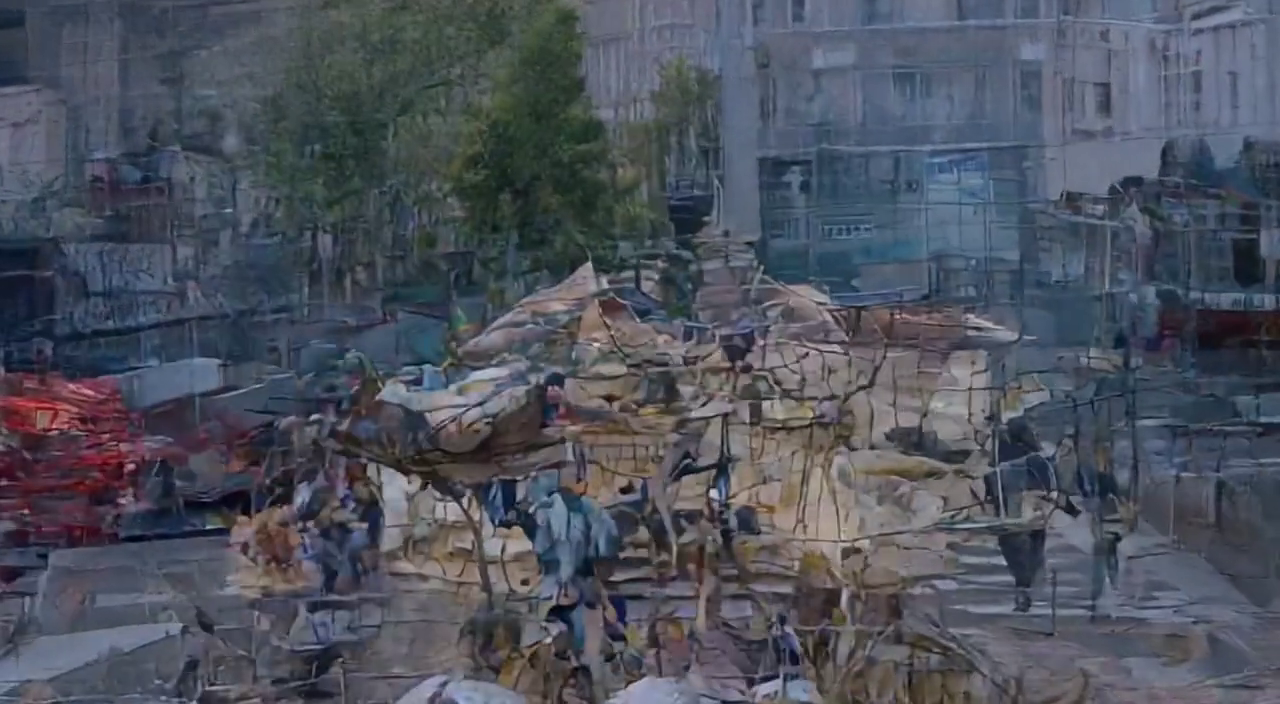}{0.3}{0.10}{0.75}{0.65}
& \imgbox{0.17\linewidth}{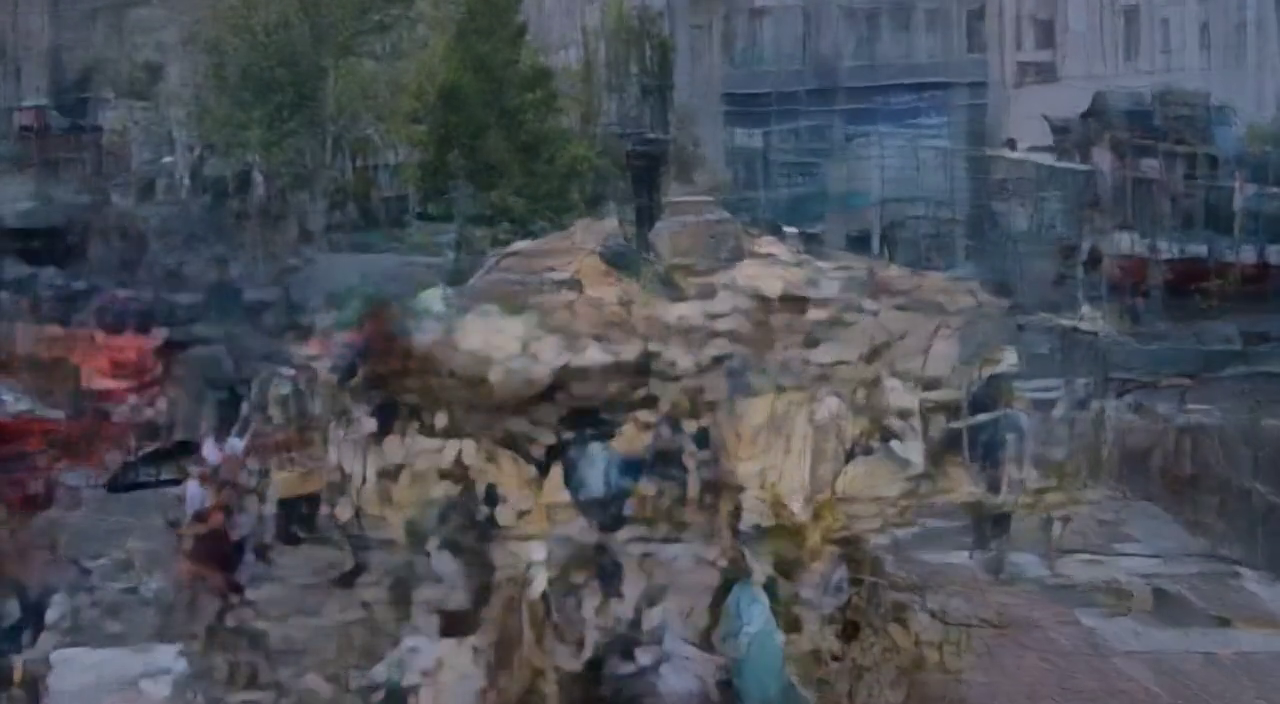}{0.3}{0.10}{0.75}{0.65}
& \imgbox{0.17\linewidth}{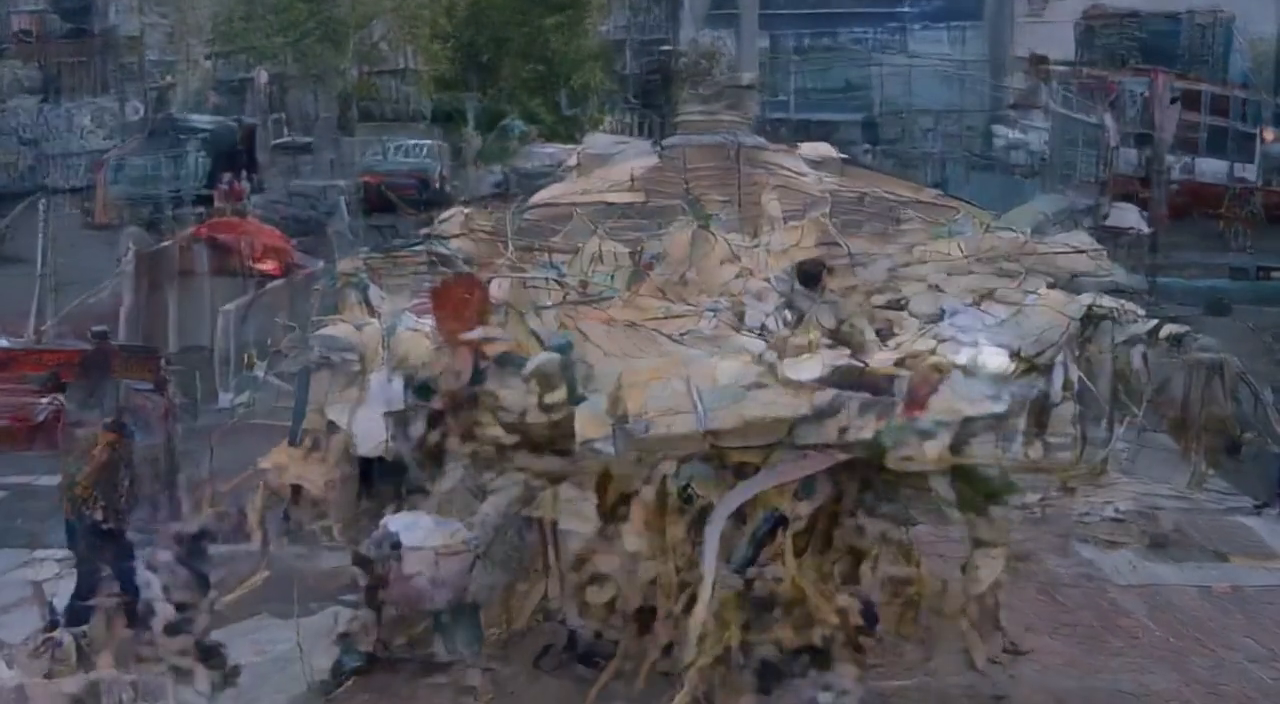}{0.3}{0.10}{0.75}{0.65}
\\

\rotatebox{90}{AW4RE}
& \includegraphics[width=0.17\linewidth]{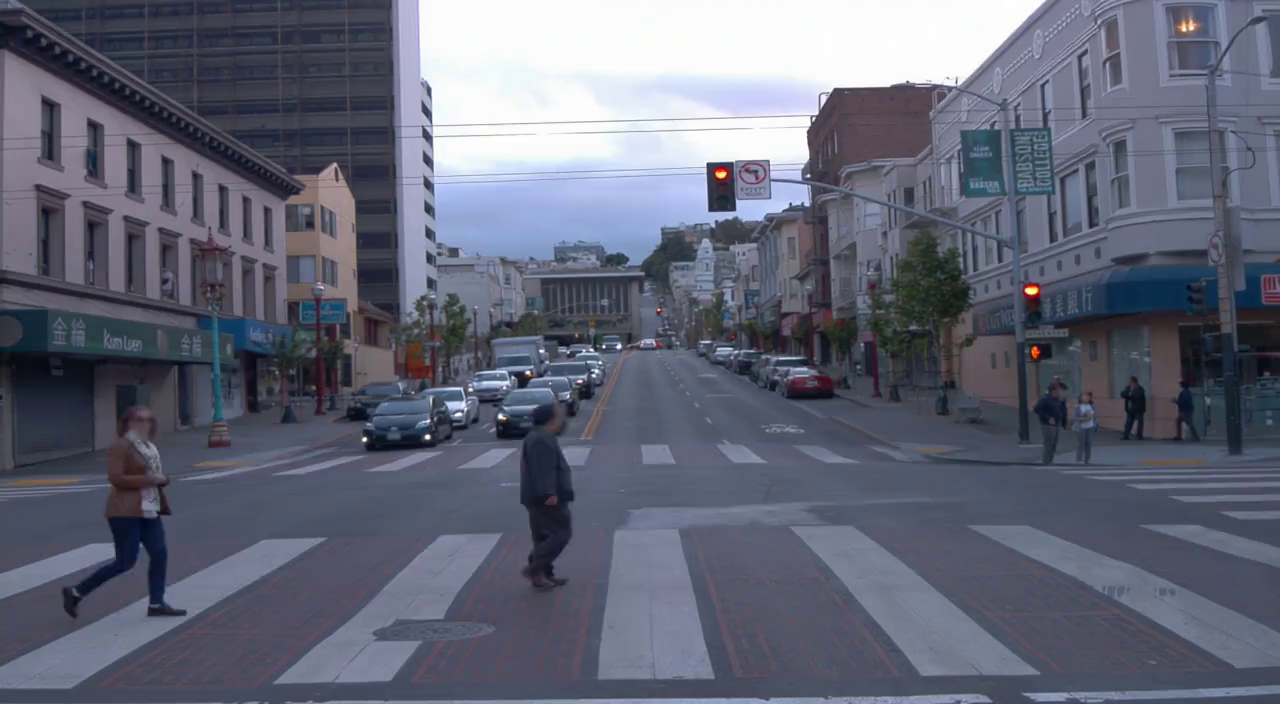}
& \imgbox{0.17\linewidth}{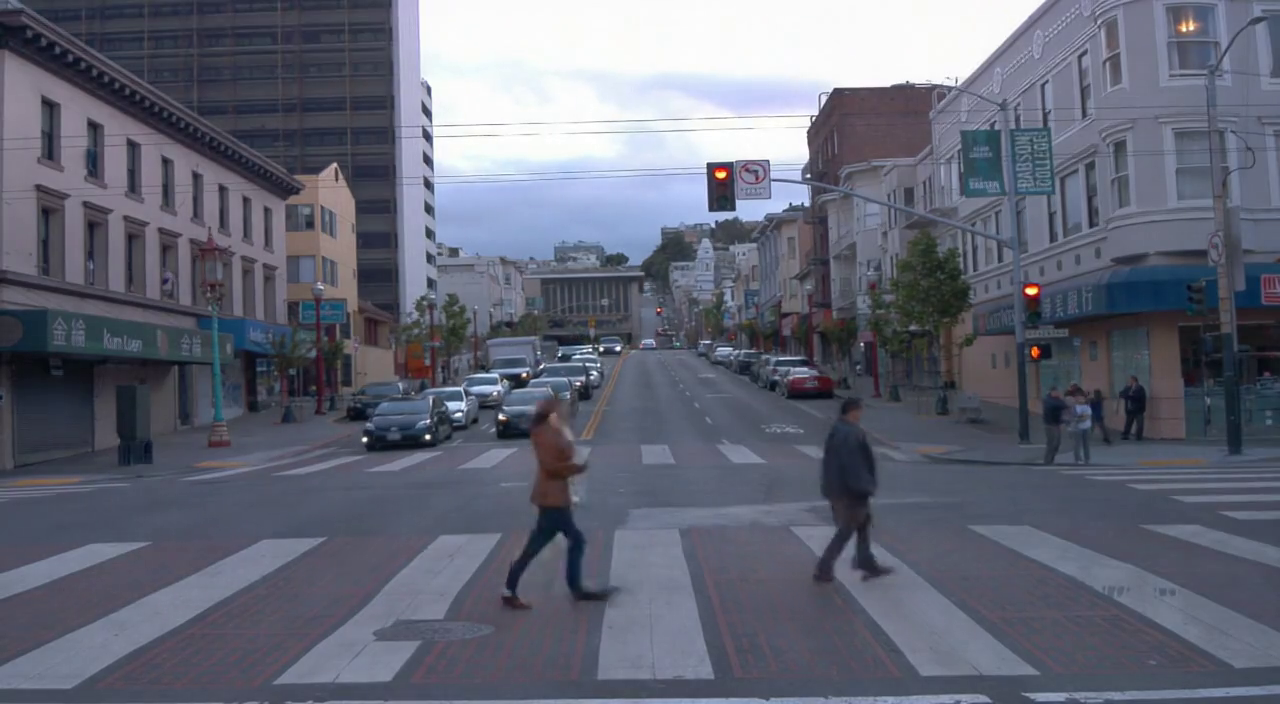}{0.35}{0.10}{0.75}{0.65}
& \imgbox{0.17\linewidth}{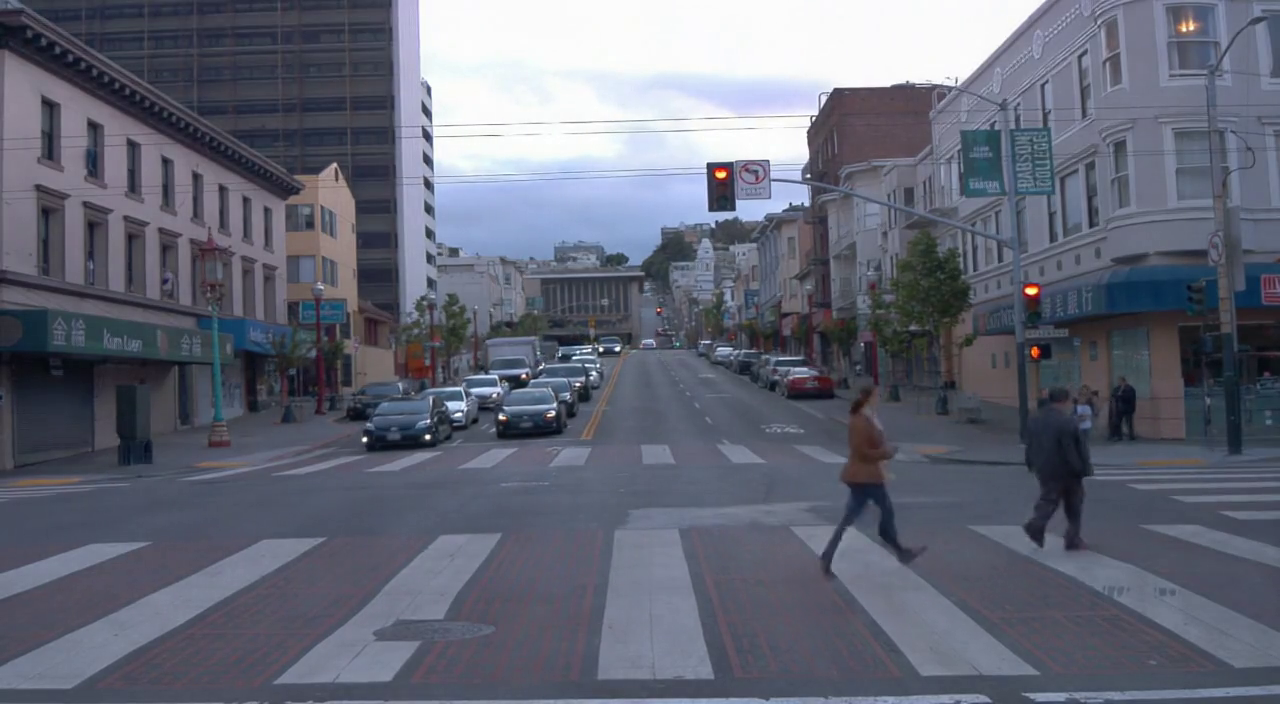}
{0.6}{0.10}{0.95}{0.65}
& \imgbox{0.17\linewidth}{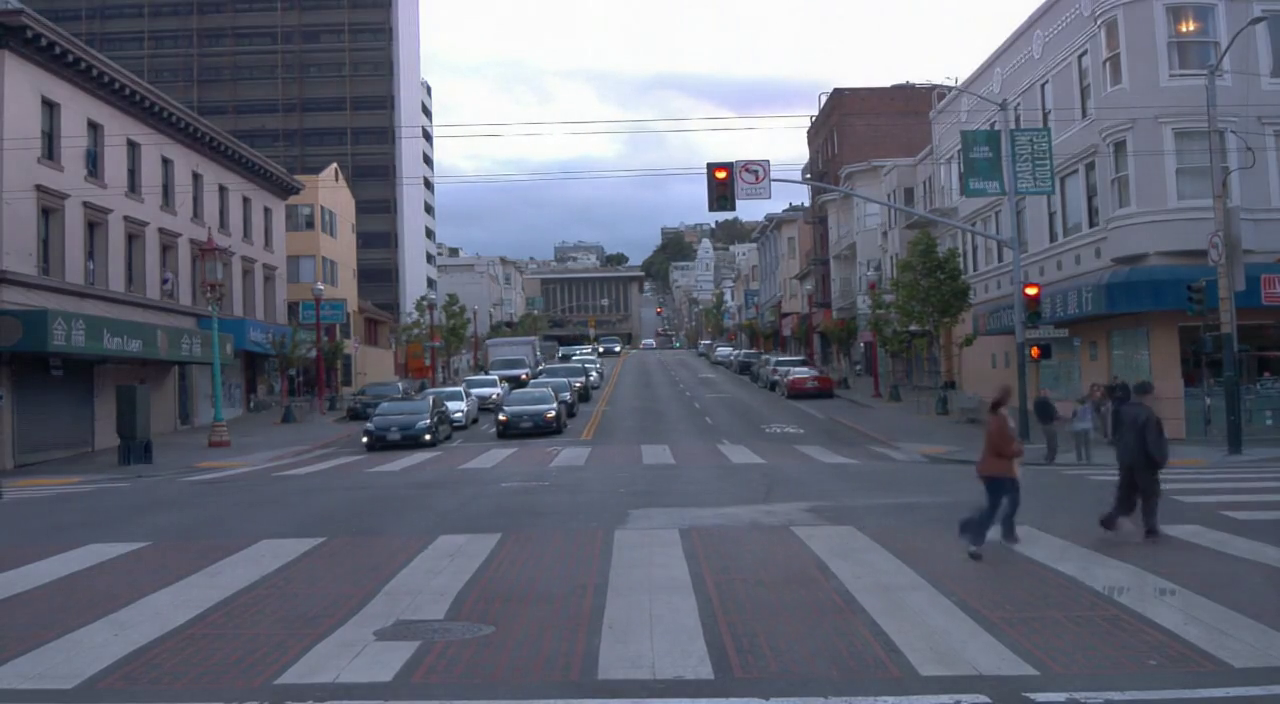}
{0.7}{0.10}{1}{0.65}
& \imgbox{0.17\linewidth}{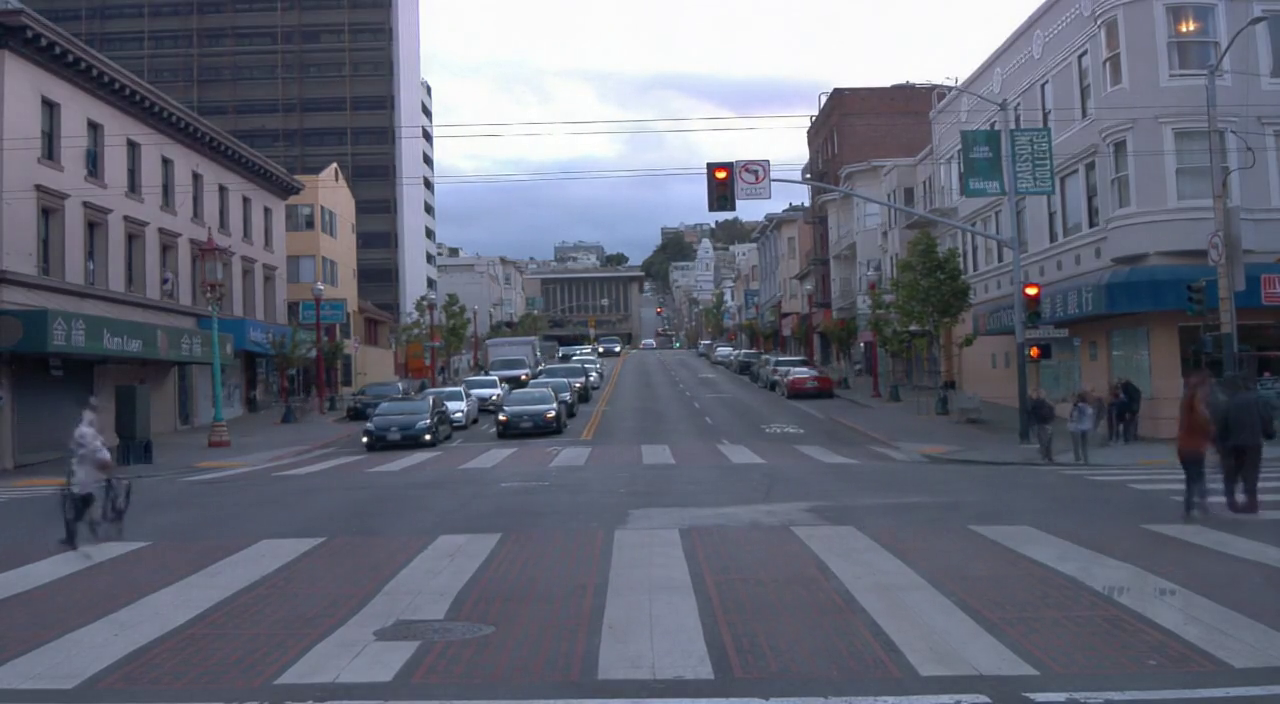}{0.9}{0.10}{1}{0.65}
\\
\end{tabular}
}

\vspace{3pt}
\caption{
Temporal Query (Next Time Steps). Top: observed evidence frames from previous iteration. Middle: observation estimates produced by GEN3C at next iteration. Bottom: observation estimates produced by AW4RE at next iteration. The 4D-informed evidence retrieval in AW4RE reduces hallucination and yields improved prediction quality. 
}
\label{fig:play_forward}
\end{figure*}

\begin{figure*}[t]
\centering
\setlength{\tabcolsep}{4pt}

{\footnotesize
\begin{tabular}{c|ccccc}
 & $o^{(1)}_1$ & $o^{(1)}_{15}$ & $o^{(1)}_{30}$ & $o^{(1)}_{55}$ & $o^{(1)}_{59}$ \\
\rotatebox{90}{Evidence}
& \includegraphics[width=0.17\linewidth]{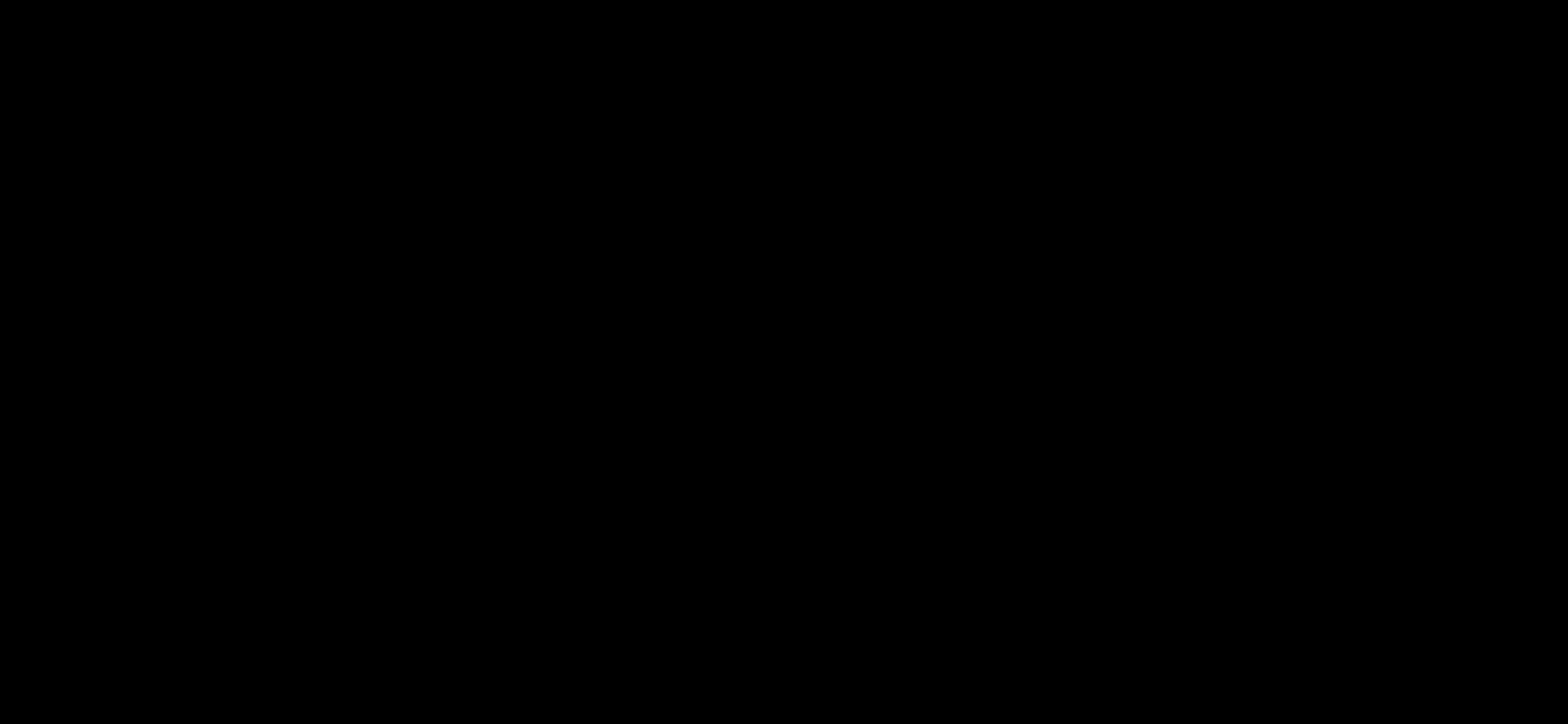}
& \includegraphics[width=0.17\linewidth]{figs/test_6/input_video/black.png}
& \includegraphics[width=0.17\linewidth]{figs/test_6/input_video/black.png}
& \includegraphics[width=0.17\linewidth]{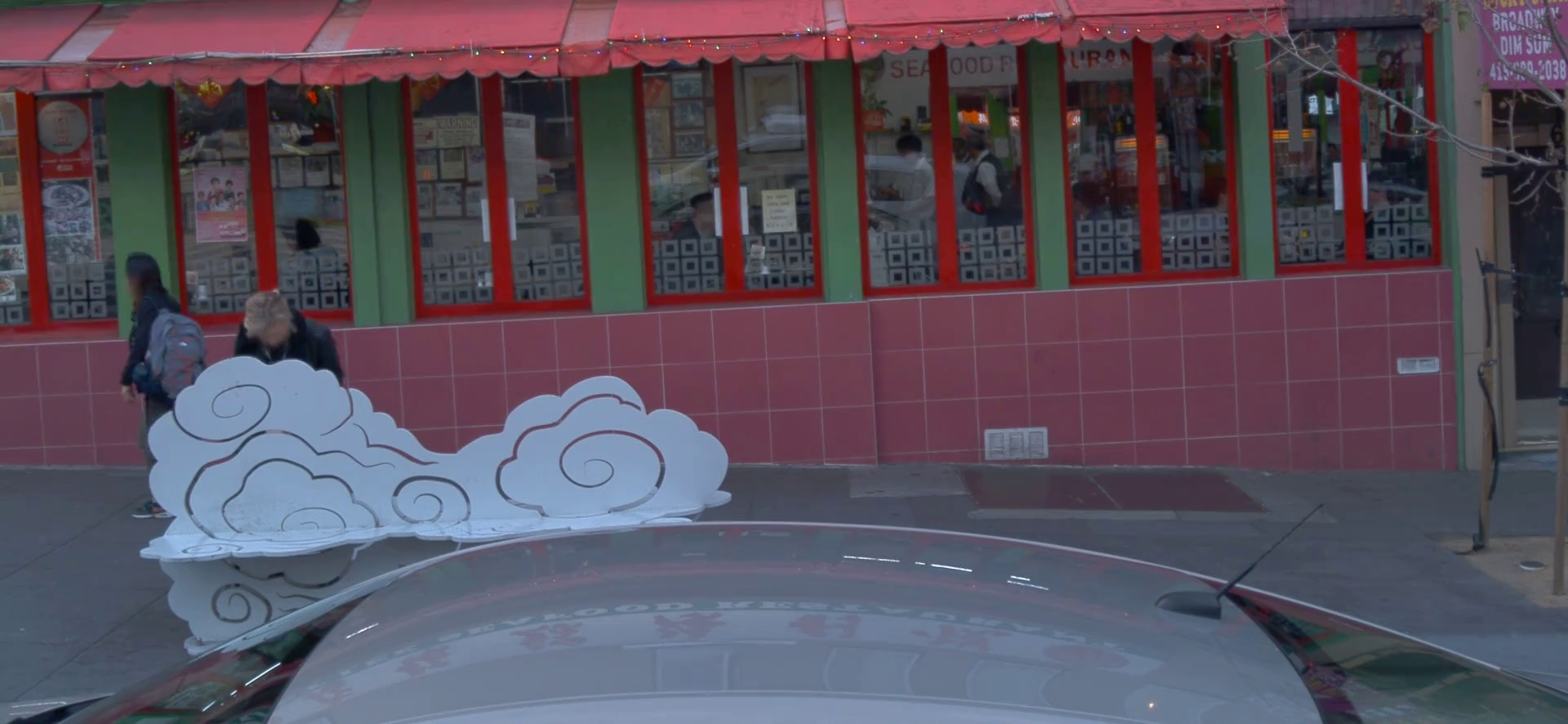}
& \includegraphics[width=0.17\linewidth]{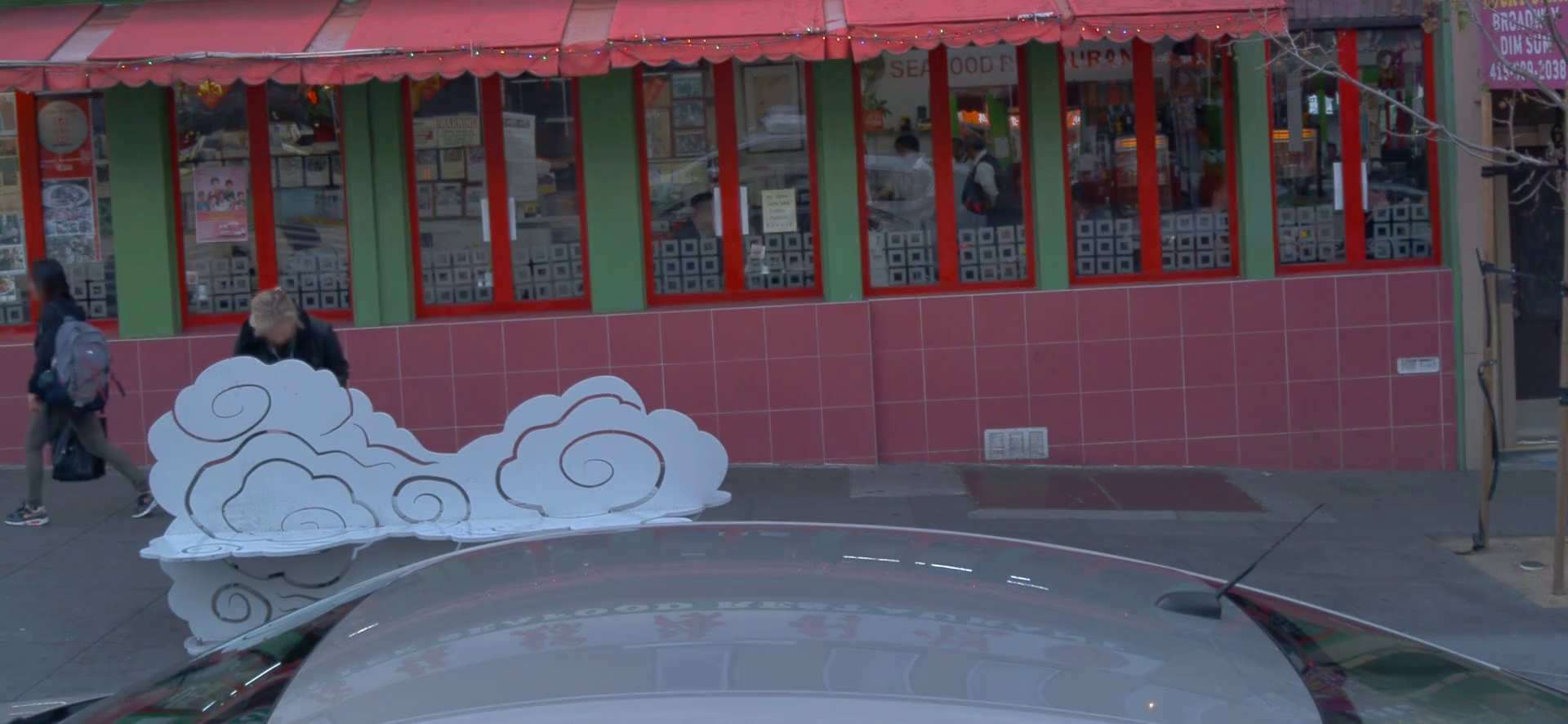}
\\
& \multicolumn{5}{c}{} \\
 & \multicolumn{5}{c}{$\mathcal{D}^{(1)} = \{(\{o^{(1)}_t\}_{t=1}^{121}, \{a^{(1)}_t\}_{t=1}^{121})\}$, \   
$a^{(2)}_t = a^{(1)}_{55} \  \forall t \in \{1,\dots,54\}, 
\quad
a^{(2)}_t = a^{(1)}_{t} \  \forall t \in \{55,\dots,121\}.
$} \\
 & \multicolumn{5}{c}{} \\
 & $o^{(2)}_1$ & $o^{(2)}_{15}$ & $o^{(2)}_{30}$ & $o^{(2)}_{55}$ & $o^{(2)}_{59}$ \\
\rotatebox{90}{GEN3C}
& \imgbox{0.17\linewidth}{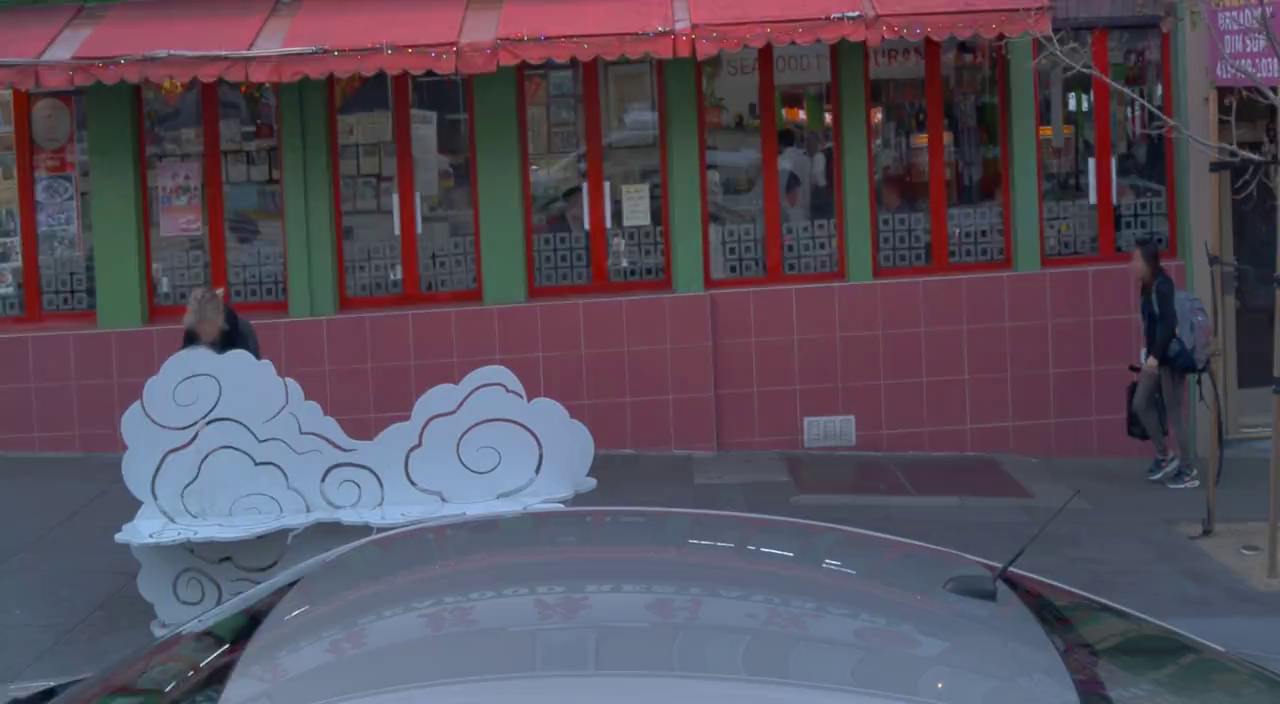}
{0.83}{0.15}{0.98}{0.75}
& \imgbox{0.17\linewidth}{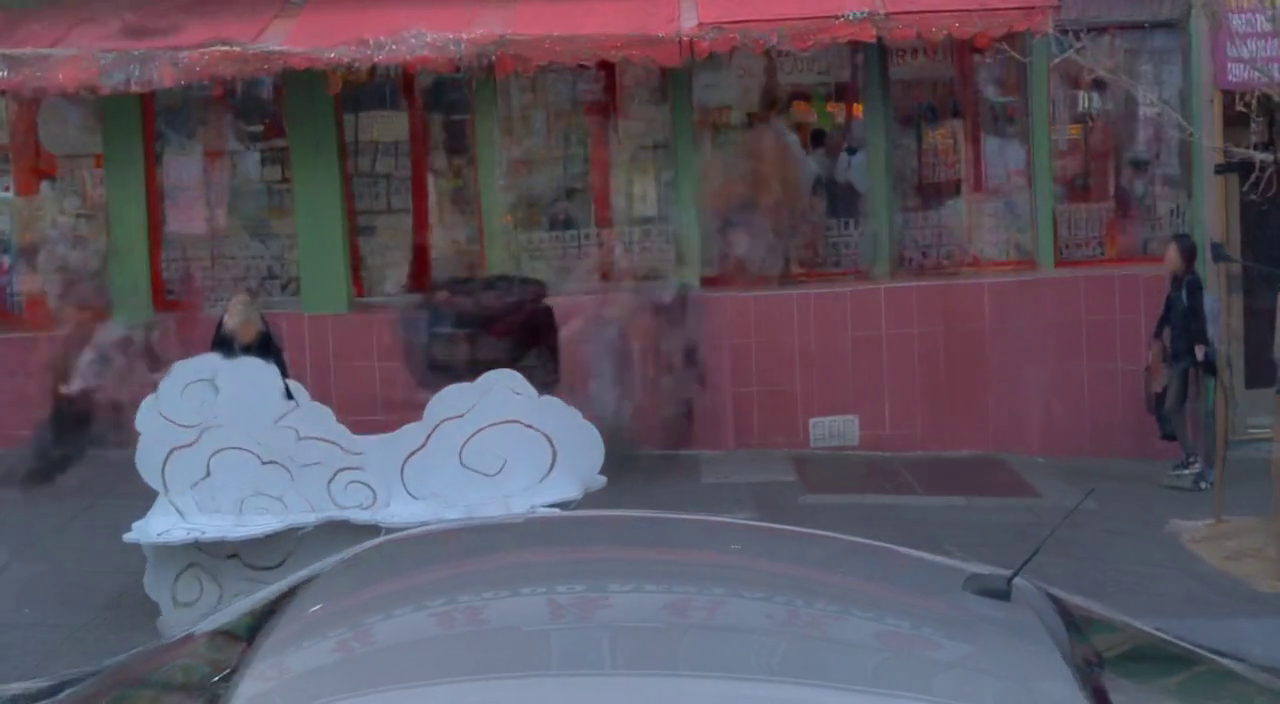}{0.83}{0.15}{0.98}{0.75}
& \imgbox{0.17\linewidth}{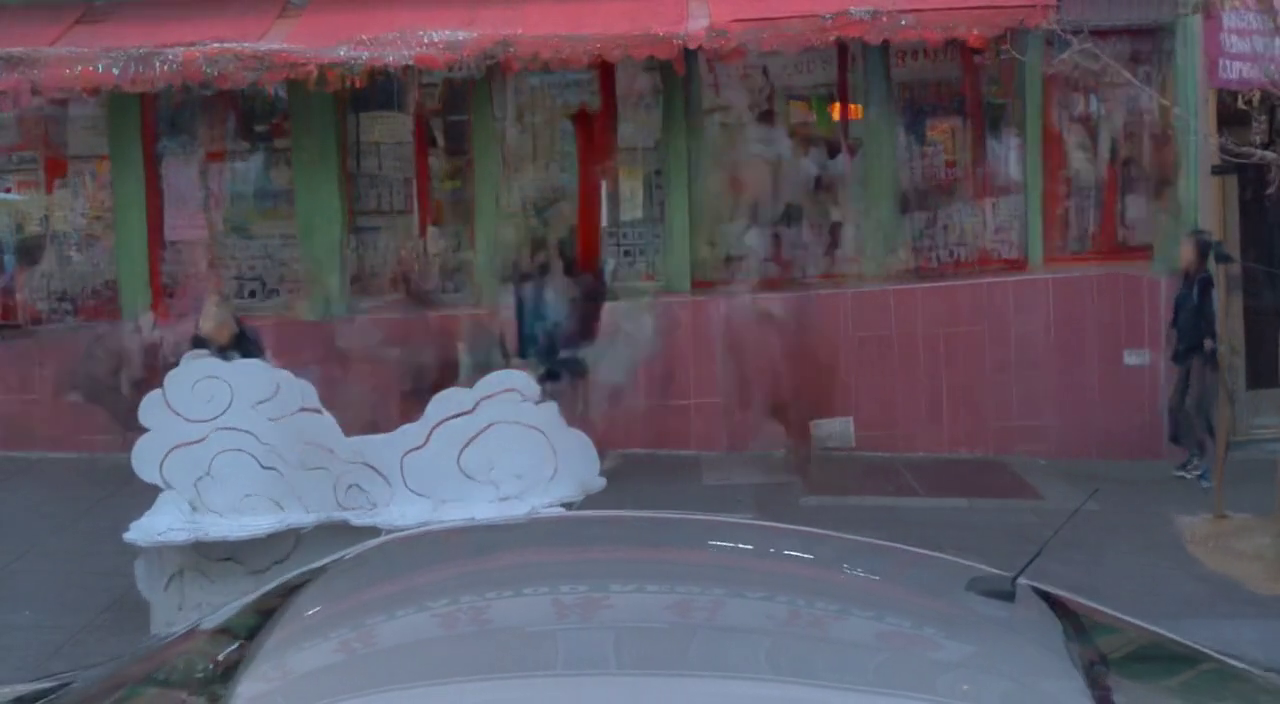}{0.83}{0.15}{0.98}{0.75}
& \imgbox{0.17\linewidth}{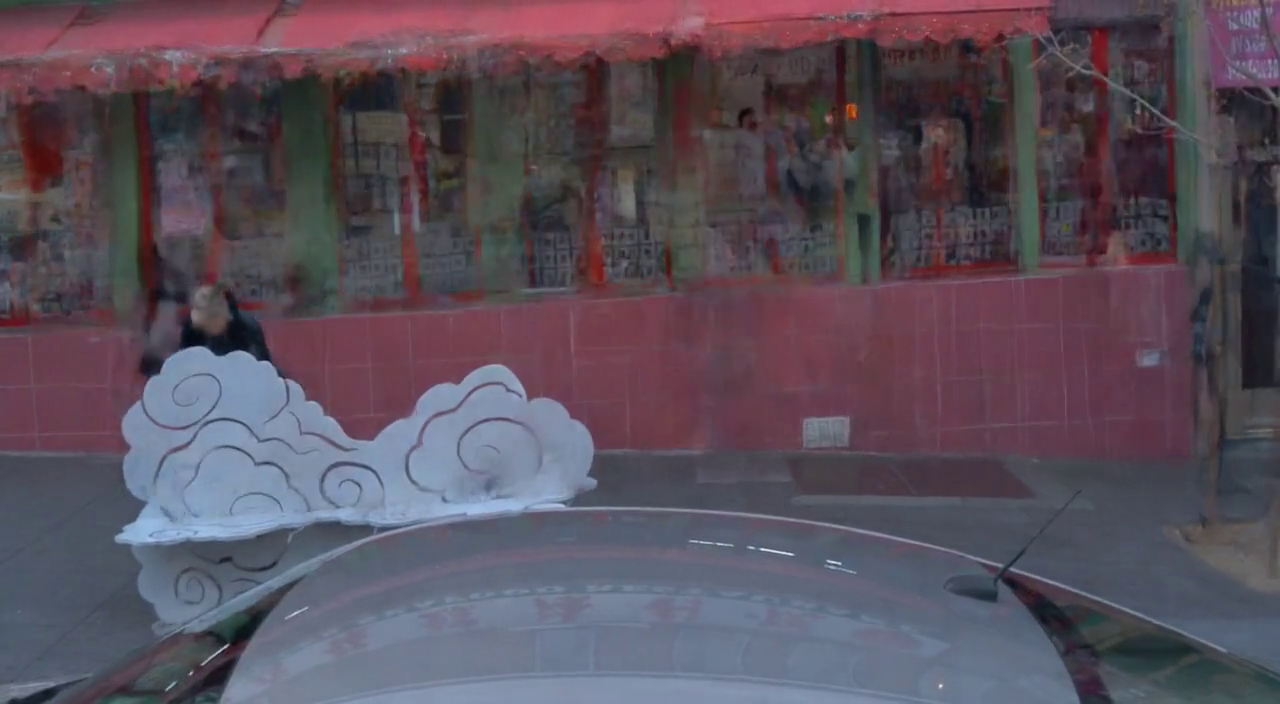}
{0.05}{0.15}{0.2}{0.75}
& \imgbox{0.17\linewidth}{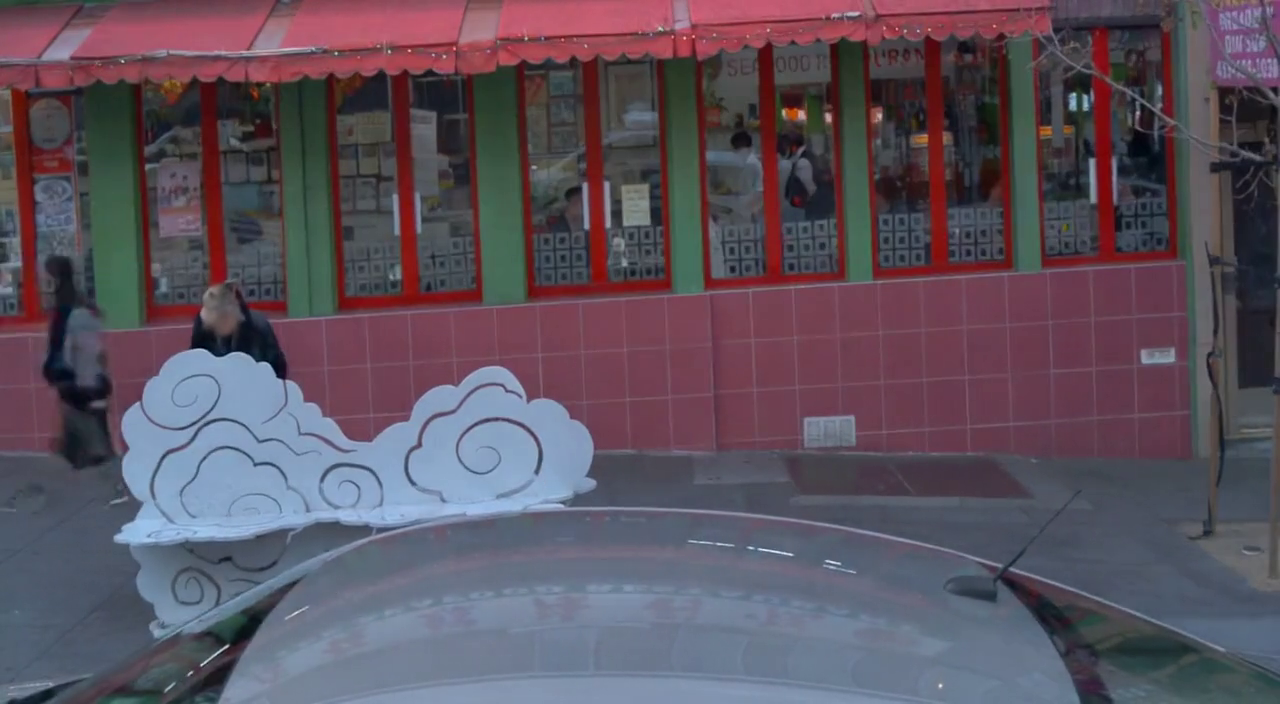}{0}{0.15}{0.15}{0.75}
\\

\rotatebox{90}{AW4RE}
& \imgbox{0.17\linewidth}{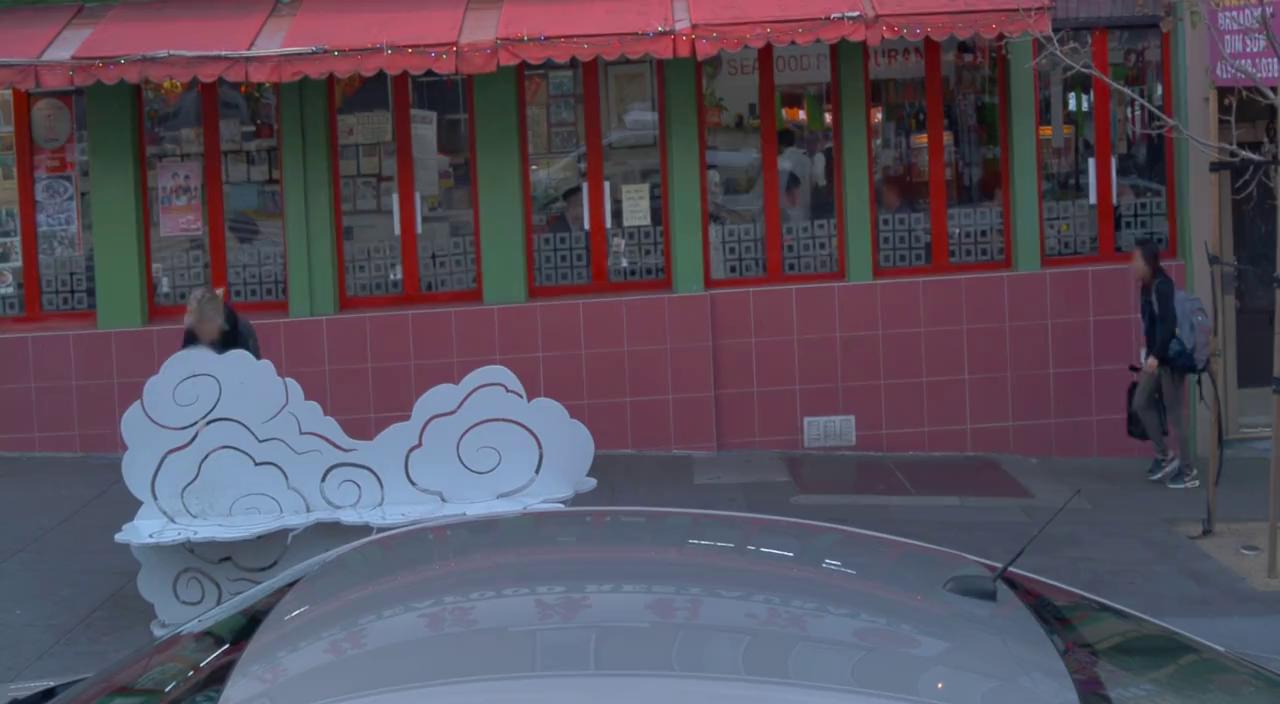}{0.83}{0.15}{0.98}{0.75}
& \imgbox{0.17\linewidth}{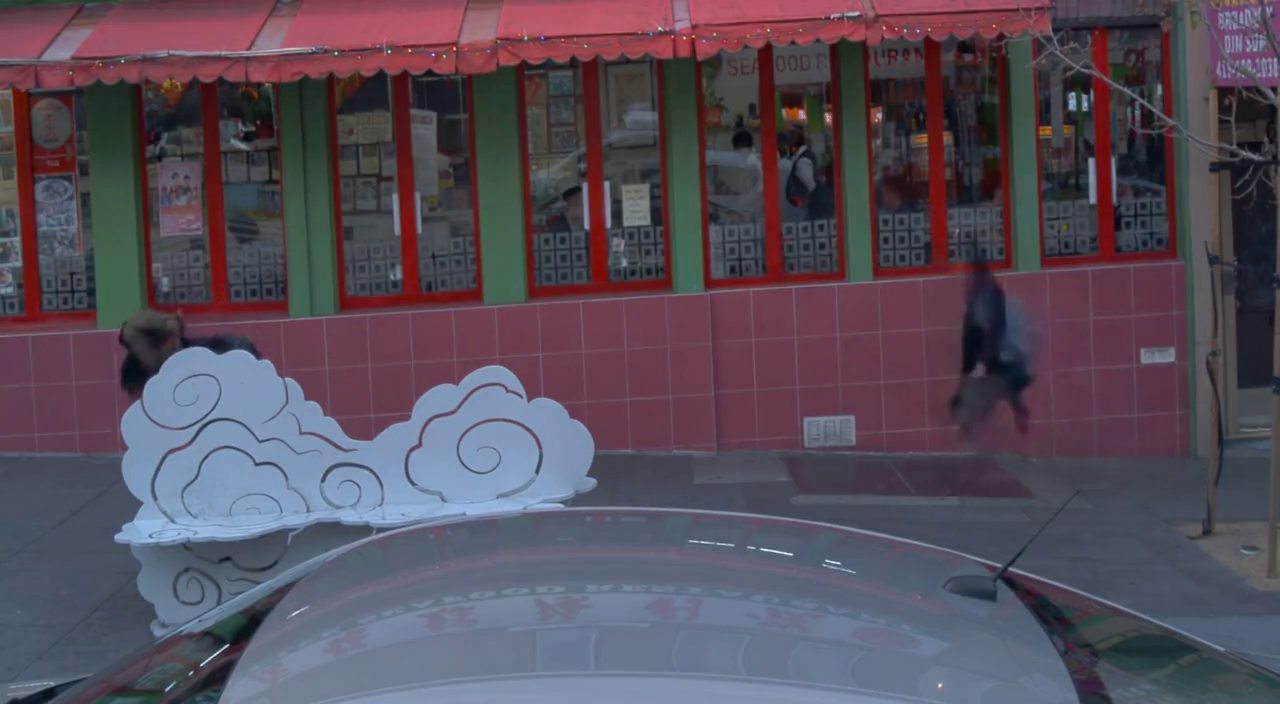}{0.7}{0.15}{0.85}{0.75}
& \imgbox{0.17\linewidth}{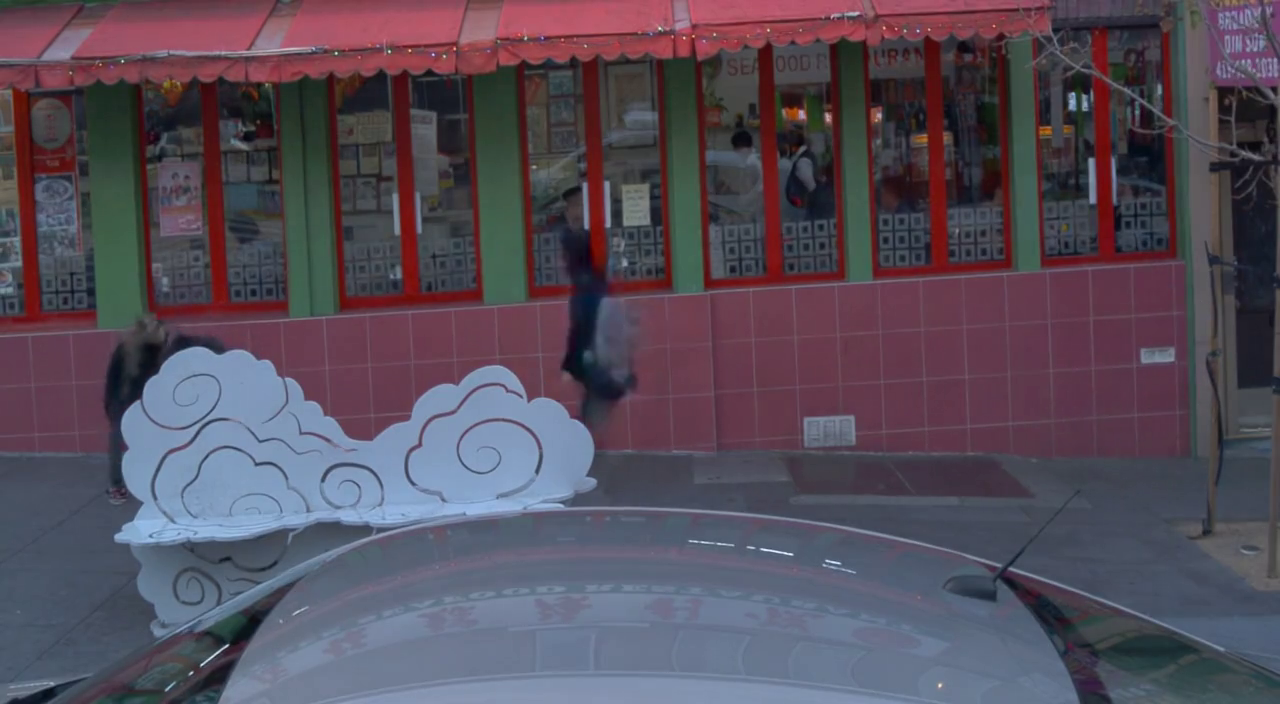}{0.4}{0.15}{0.55}{0.75}
& \imgbox{0.17\linewidth}{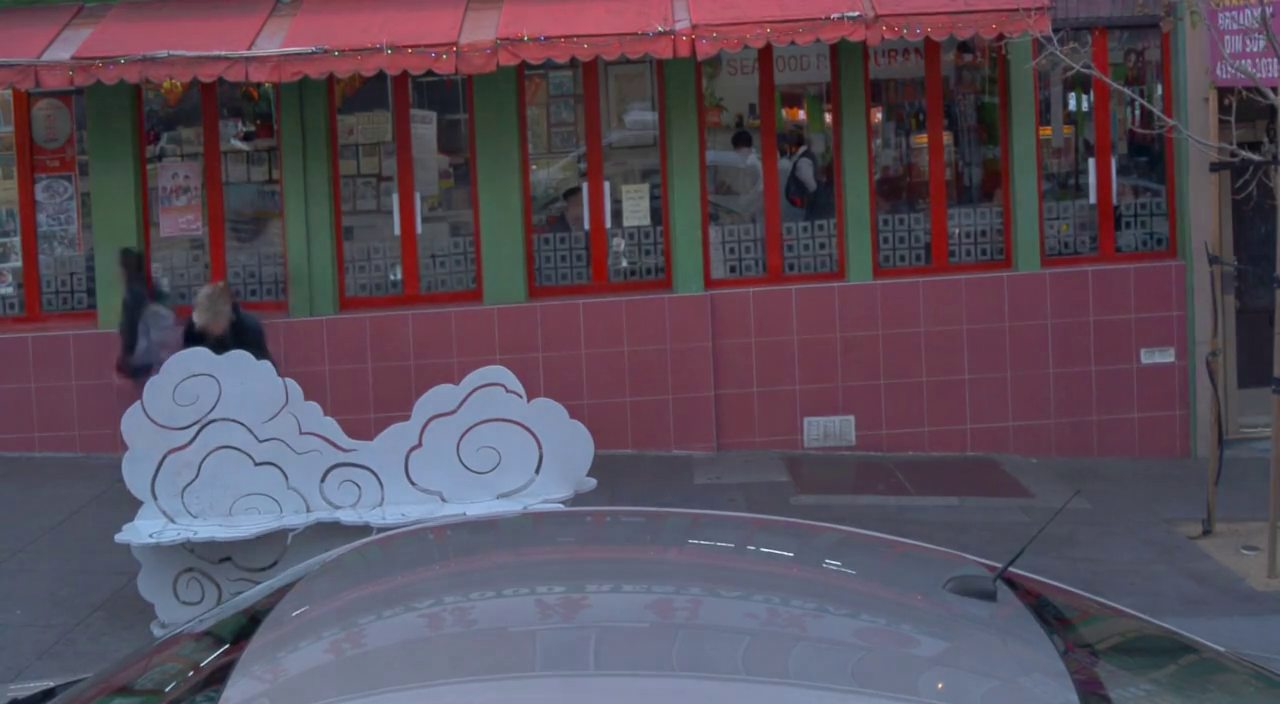}{0.05}{0.15}{0.2}{0.75}
& \imgbox{0.17\linewidth}{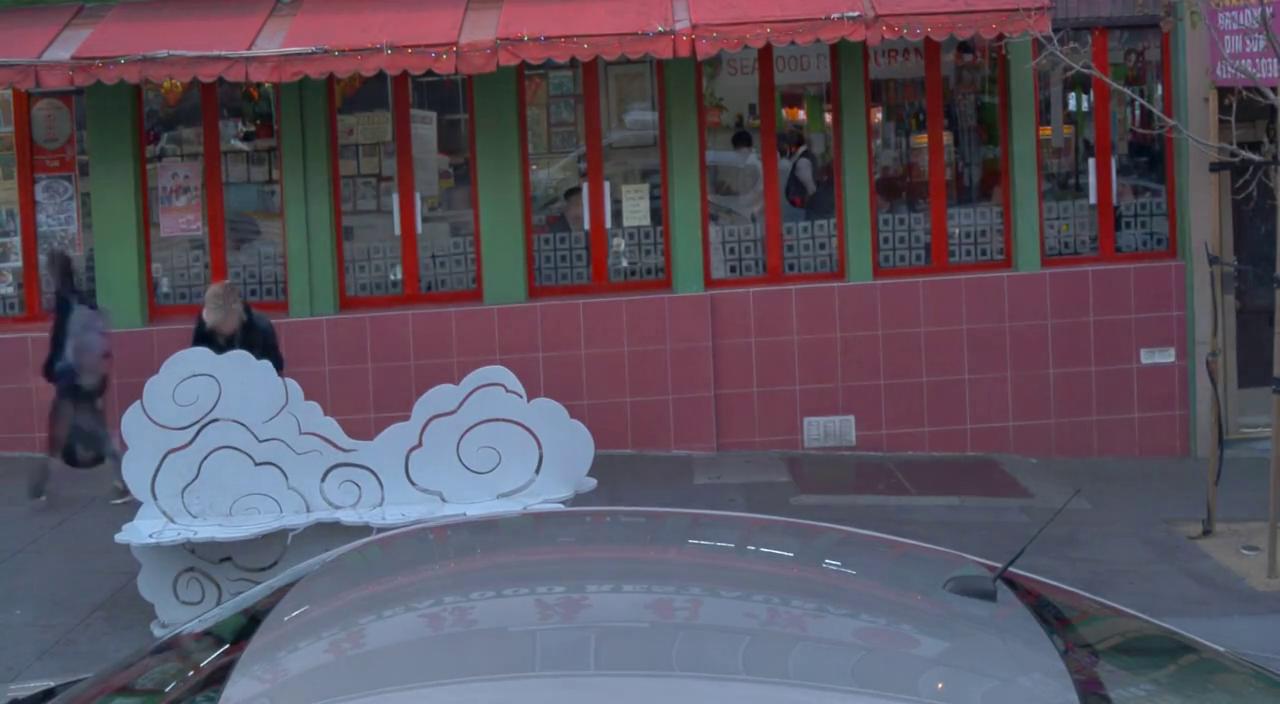}{0}{0.15}{0.15}{0.75}
\\
\end{tabular}
}

\vspace{-3pt}
\caption{
Temporal Query (Previous Time Steps). Top: observed evidence frames from previous iteration. Middle: observation estimates produced by GEN3C at next iteration. Bottom: observation estimates produced by AW4RE at next iteration. By leveraging 4D-informed evidence retrieval, AW4RE more accurately tracks dynamic objects across time. 
}
\label{fig:interwind}
\end{figure*}

\begin{figure*}[t]
\centering
\setlength{\tabcolsep}{4pt}

{\footnotesize
\begin{tabular}{c|ccccc}
 & $o^{(1)}_1$ & $o^{(1)}_{20}$ & $o^{(1)}_{40}$ & $o^{(1)}_{50}$ & $o^{(1)}_{70}$ \\
\rotatebox{90}{Evidence}
& \imgbox{0.17\linewidth}{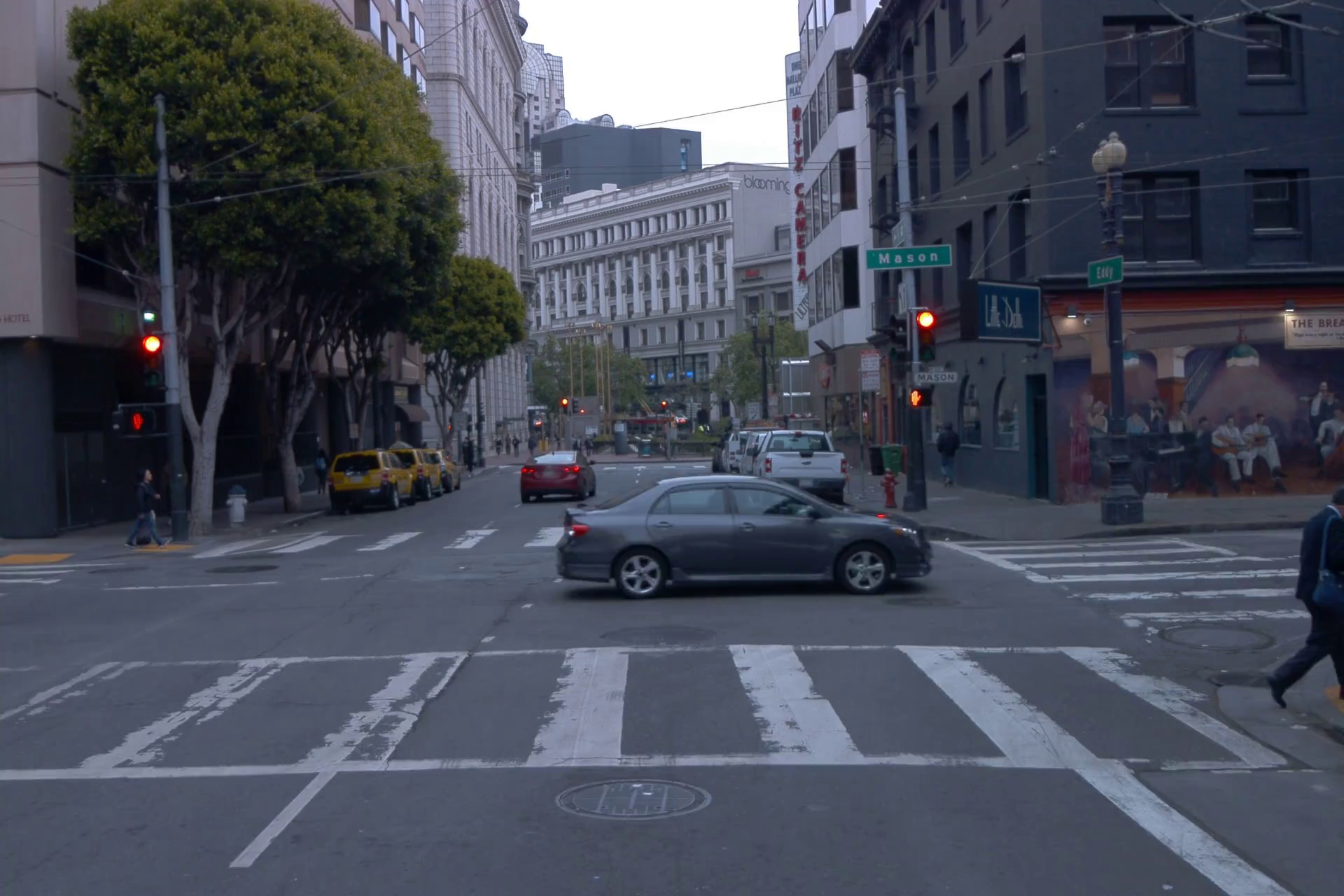}{0.35}{0.37}{0.55}{0.55}
& \imgbox{0.17\linewidth}{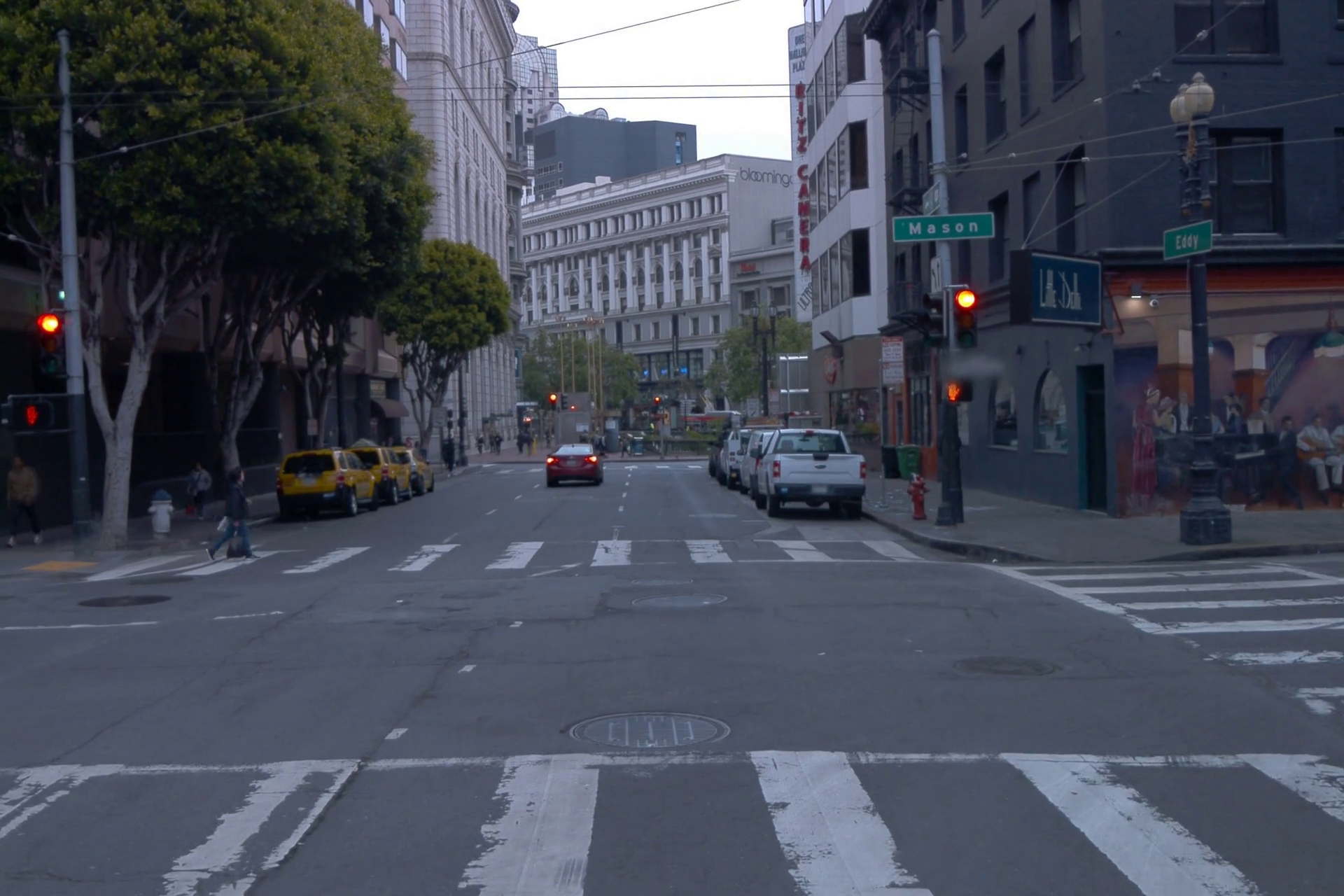}{0.35}{0.37}{0.55}{0.55}
&\imgbox{0.17\linewidth}{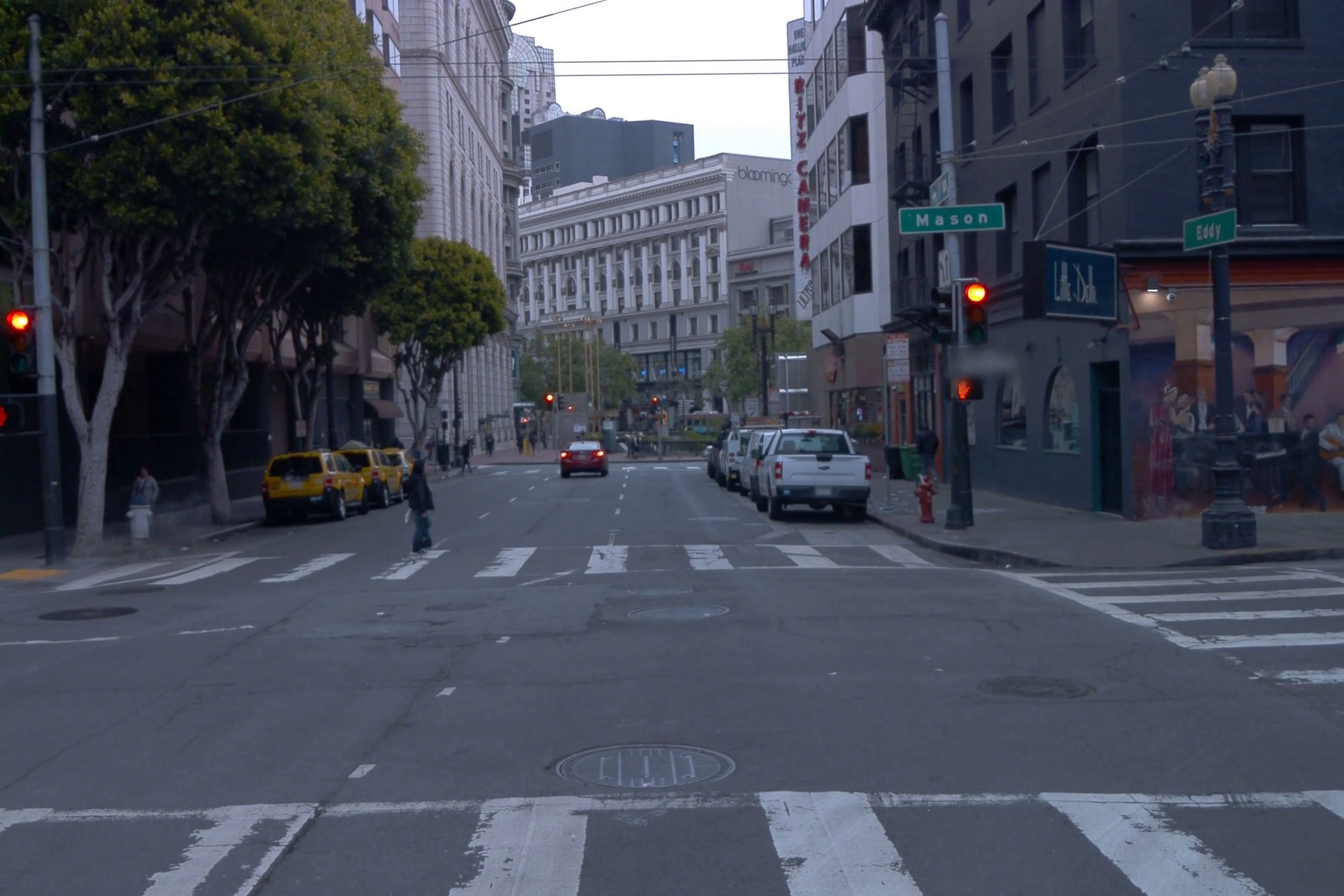}{0.28}{0.37}{0.48}{0.55}
& \imgbox{0.17\linewidth}{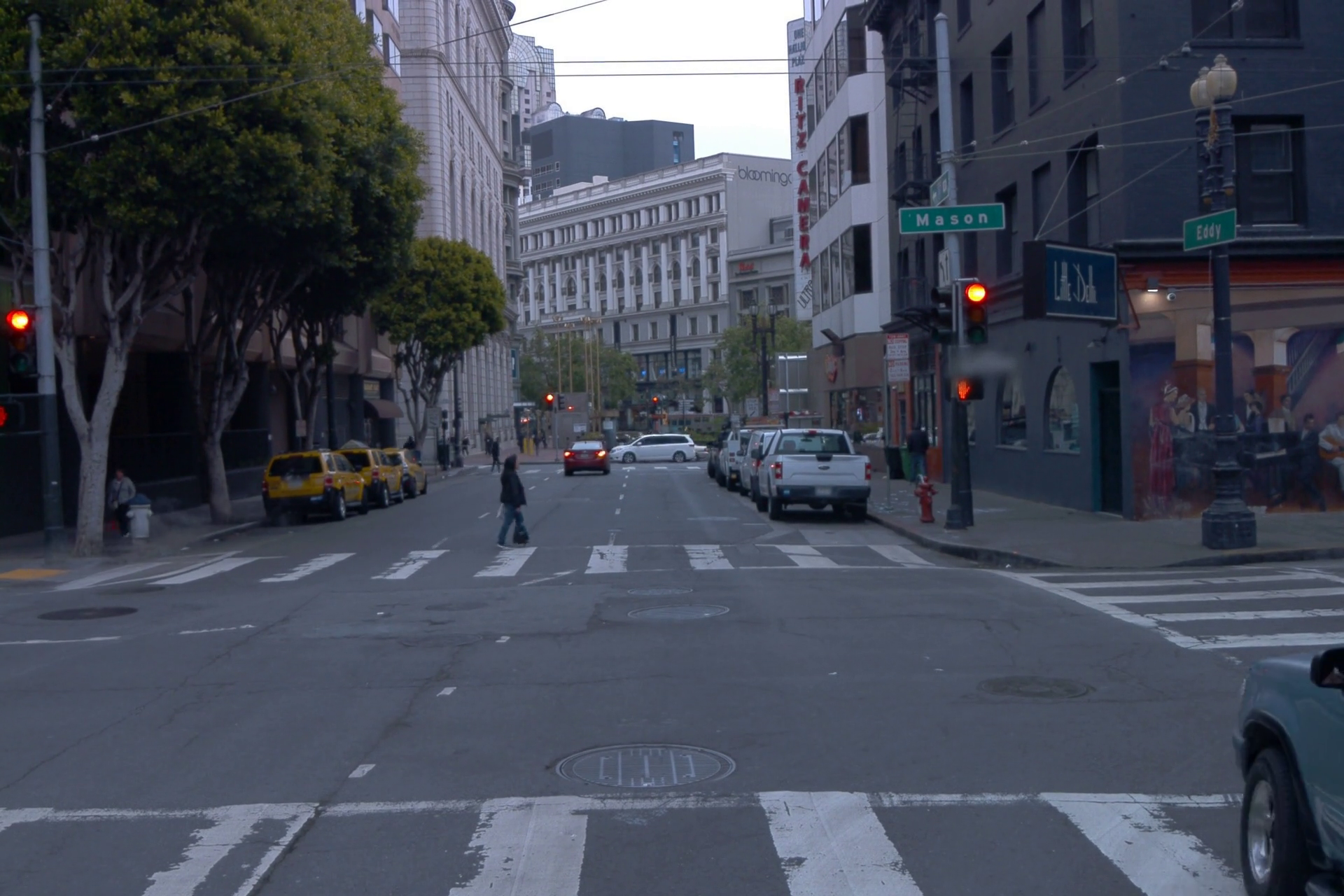}{0.35}{0.37}{0.55}{0.55}
& \imgbox{0.17\linewidth}{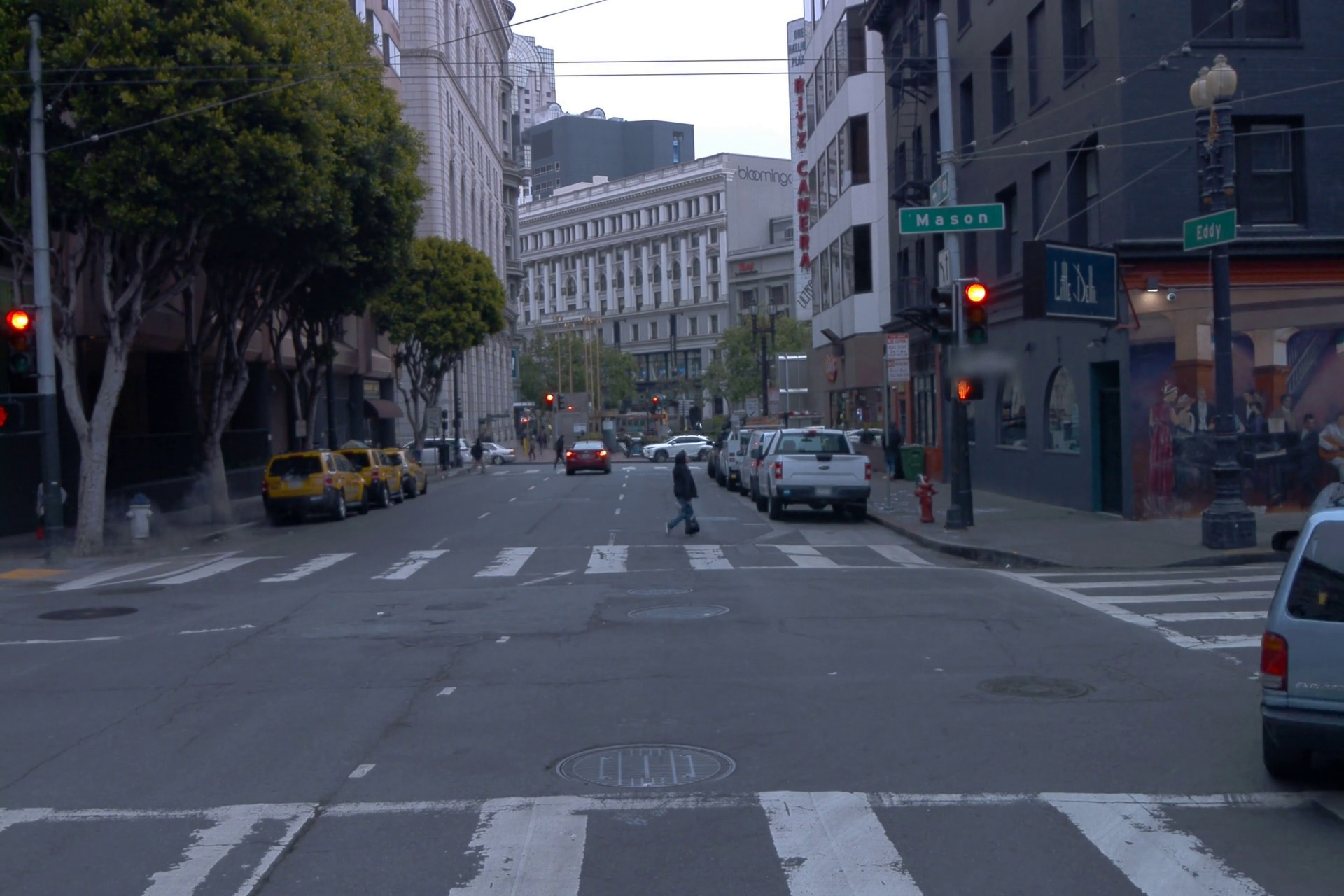}{0.4}{0.37}{0.6}{0.55}
\\
& \multicolumn{5}{c}{} \\
 & \multicolumn{5}{c}{$\mathcal{D}^{(1)} = \{(\{o^{(1)}_t\}_{t=1}^{121}, \{a^{(1)}_t\}_{t=1}^{121})\}$, $a^{(2)} = \textit{zoomed-in camera trajectory}$} \\
 & \multicolumn{5}{c}{} \\
 & $o^{(2)}_1$ & $o^{(2)}_{20}$ & $o^{(2)}_{40}$ & $o^{(2)}_{50}$ & $o^{(2)}_{70}$ \\
\rotatebox{90}{GEN3C}
&\imgbox{0.17\linewidth}{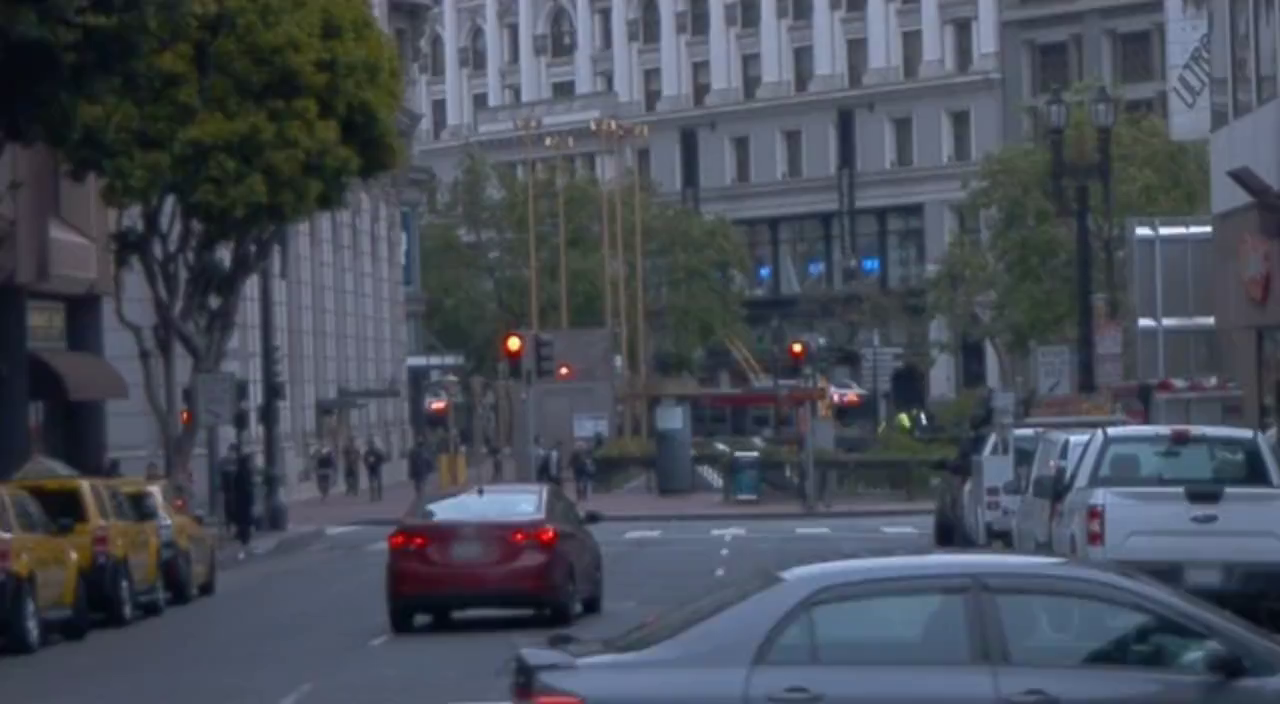}{0.15}{0}{0.75}{0.55}
& \imgbox{0.17\linewidth}{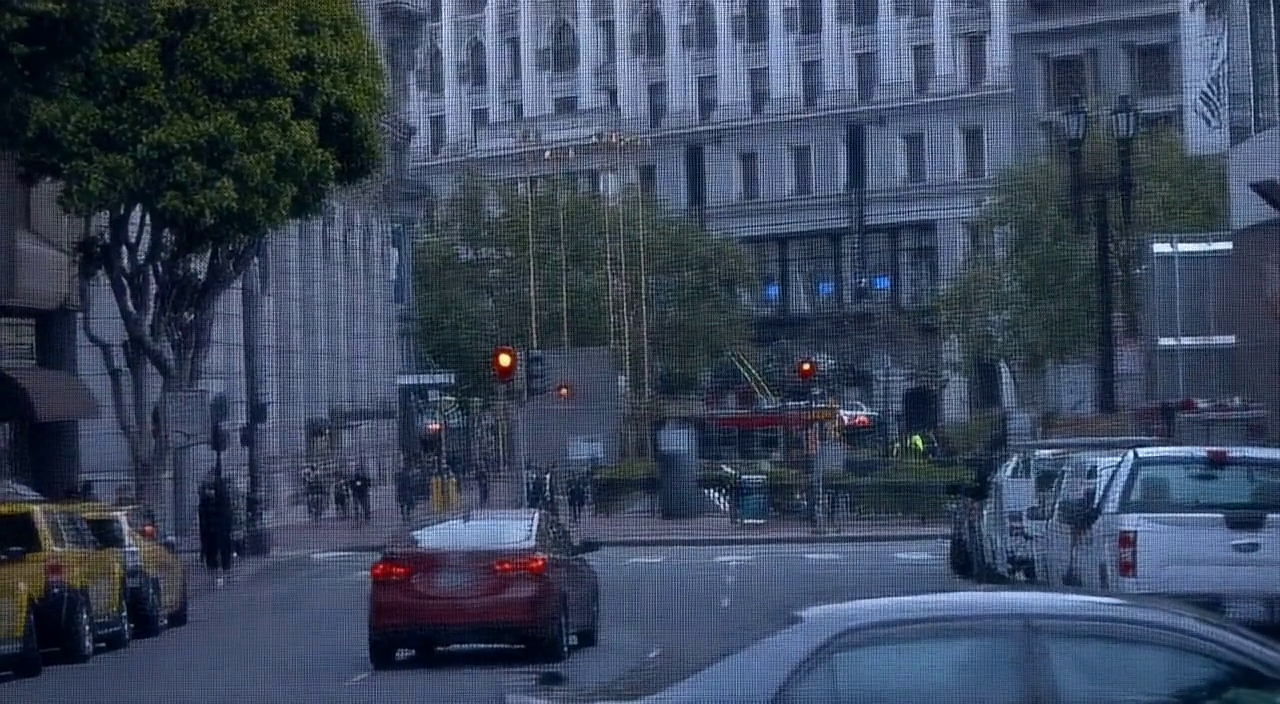}{0.15}{0}{0.75}{0.55}
& \imgbox{0.17\linewidth}{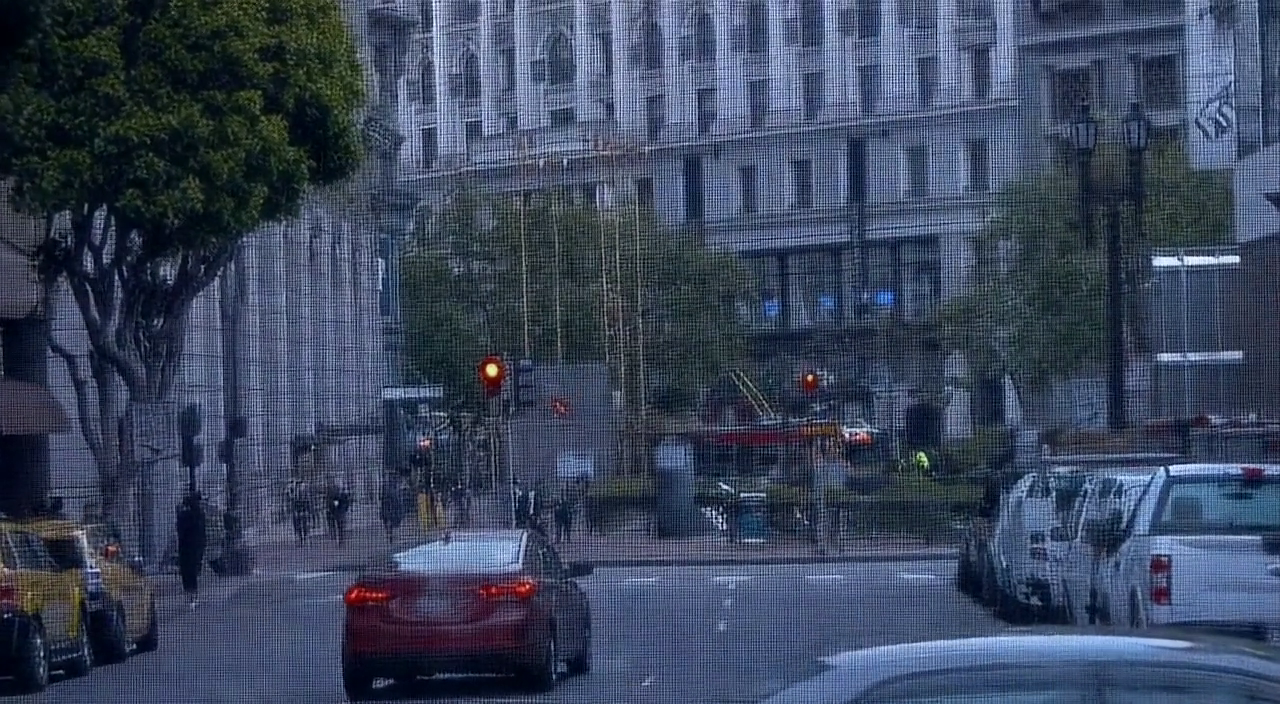}{0}{0}{0.60}{0.55}
& \imgbox{0.17\linewidth}{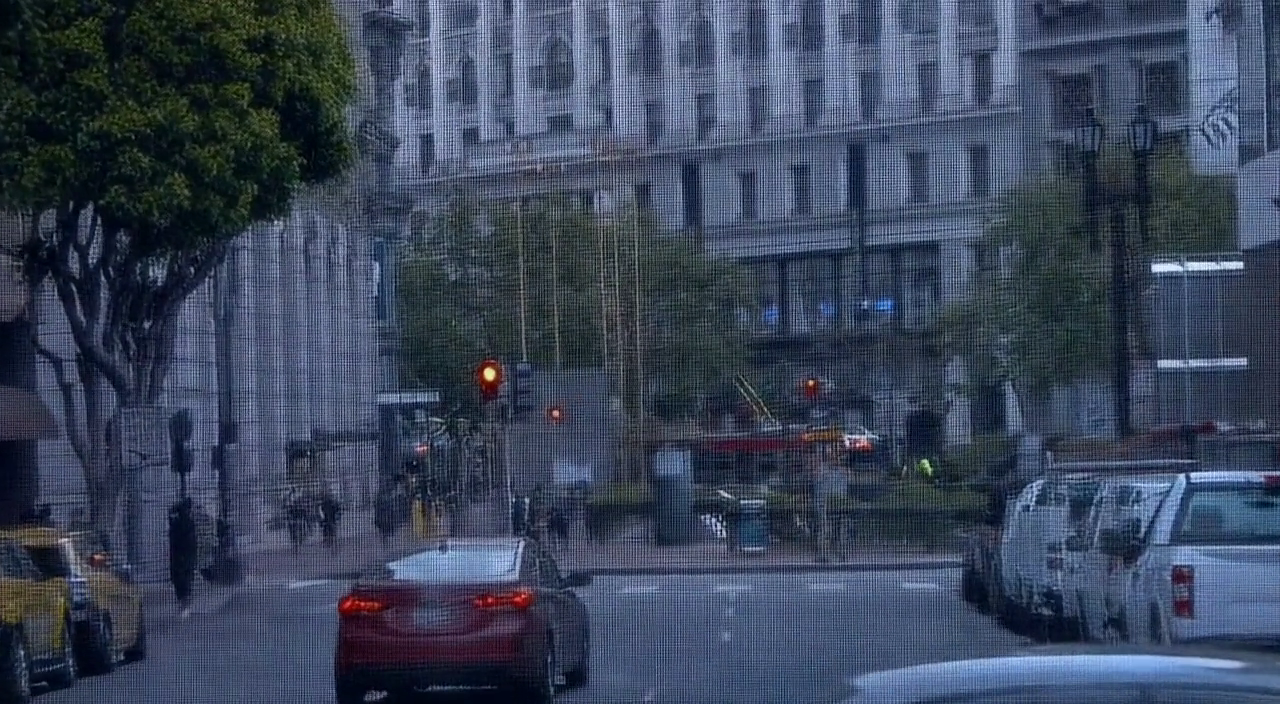}{0.15}{0}{0.75}{0.55}
&\imgbox{0.17\linewidth}{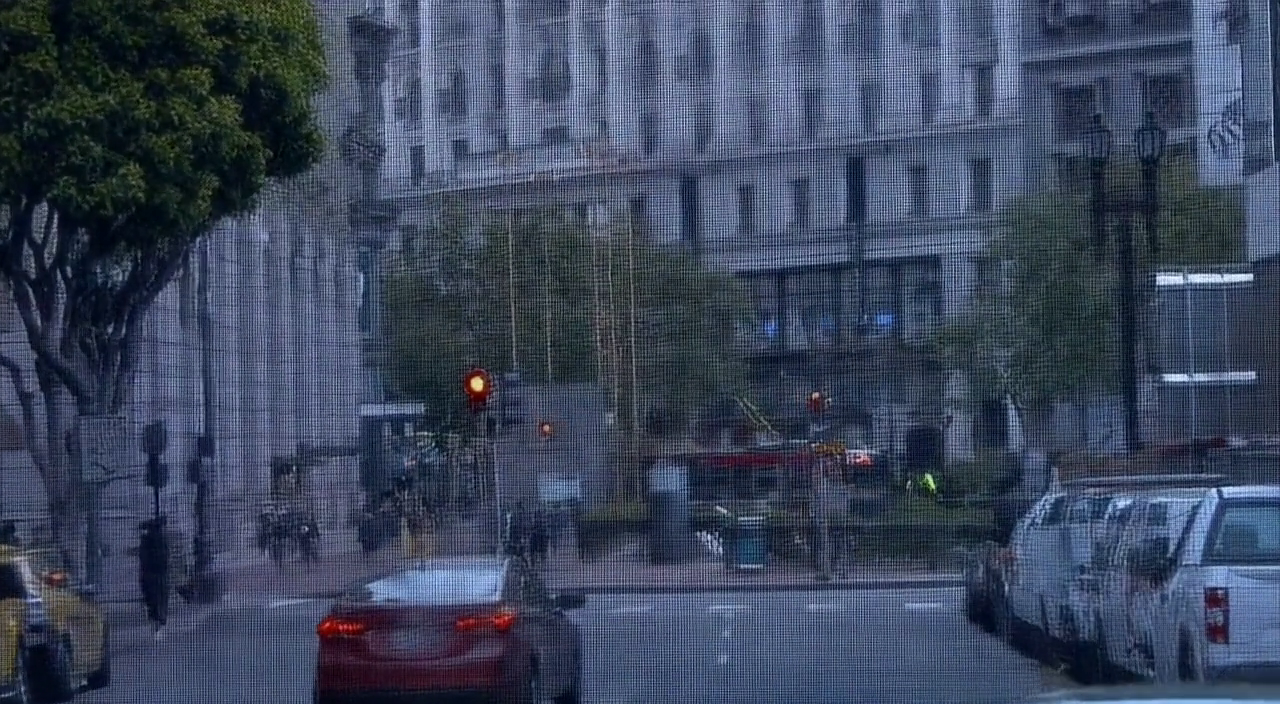}{0.15}{0}{0.75}{0.55}
\\

\rotatebox{90}{AW4RE}
& \imgbox{0.17\linewidth}{figs/test_3/kavai/test_3_ours_frame_0001.png}{0.15}{0}{0.75}{0.55}
& \imgbox{0.17\linewidth}{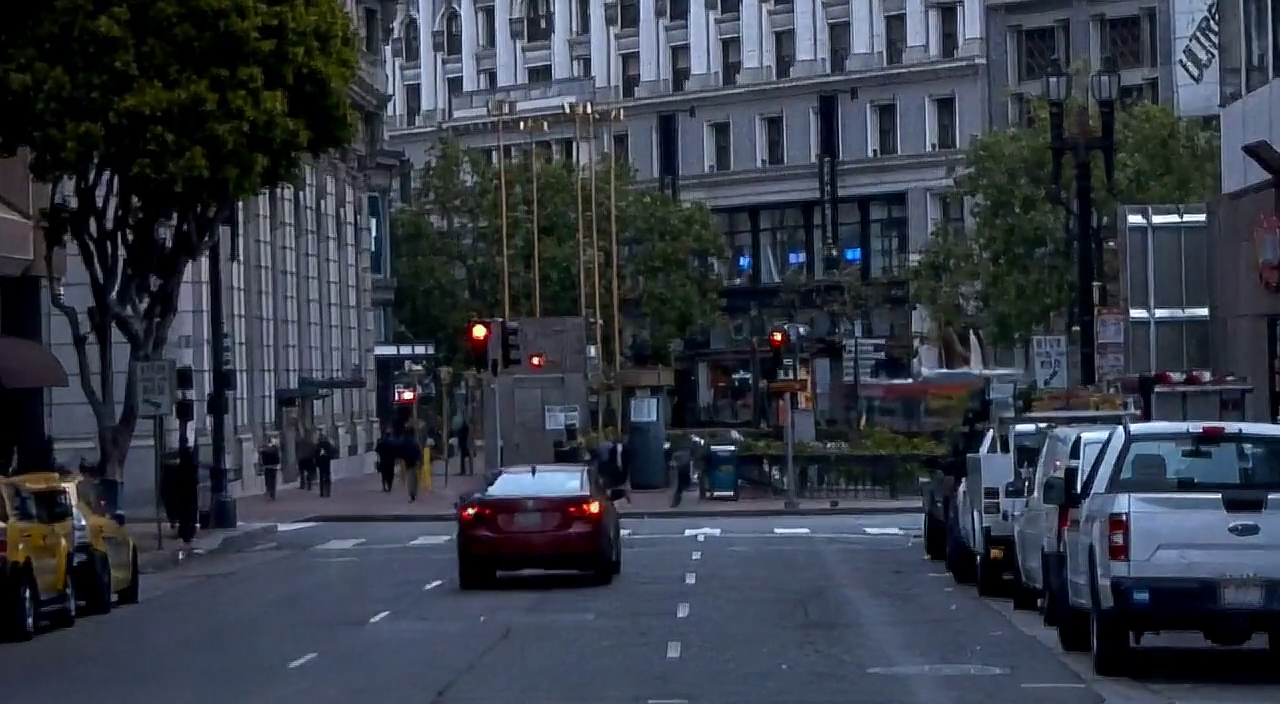}{0.15}{0}{0.75}{0.55}
& \imgbox{0.17\linewidth}{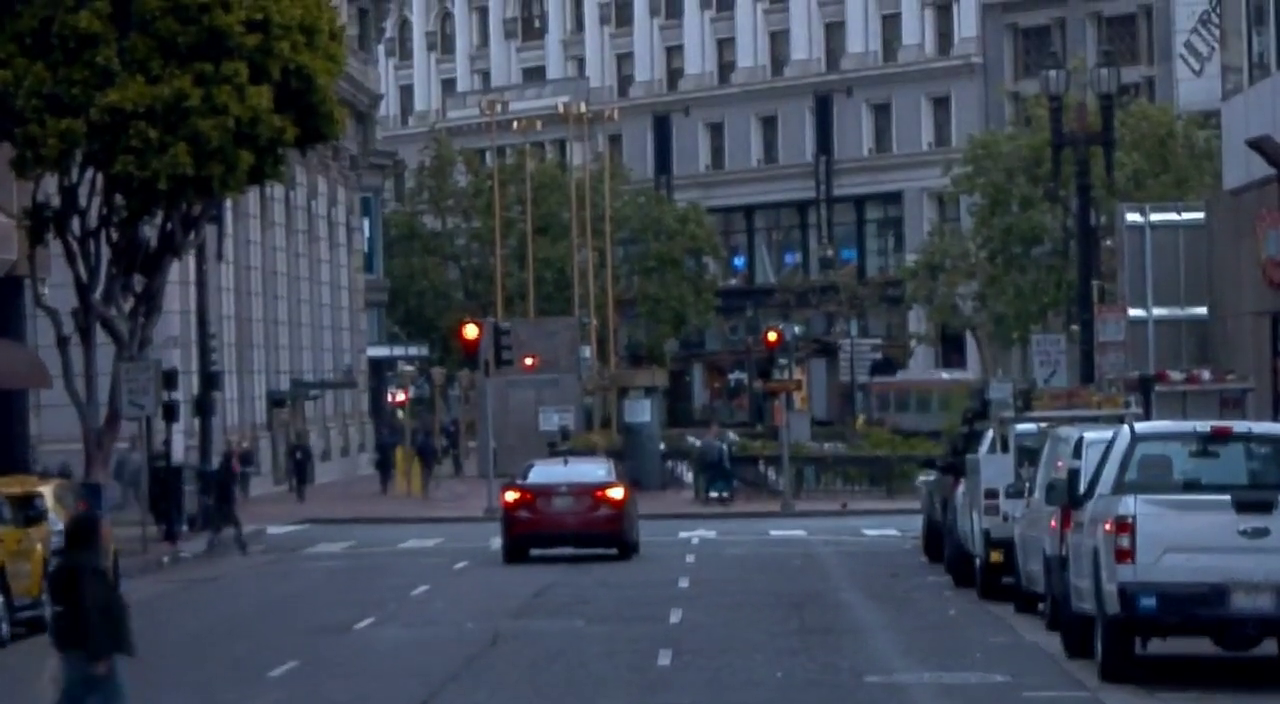}{0}{0}{0.60}{0.55}
& \imgbox{0.17\linewidth}{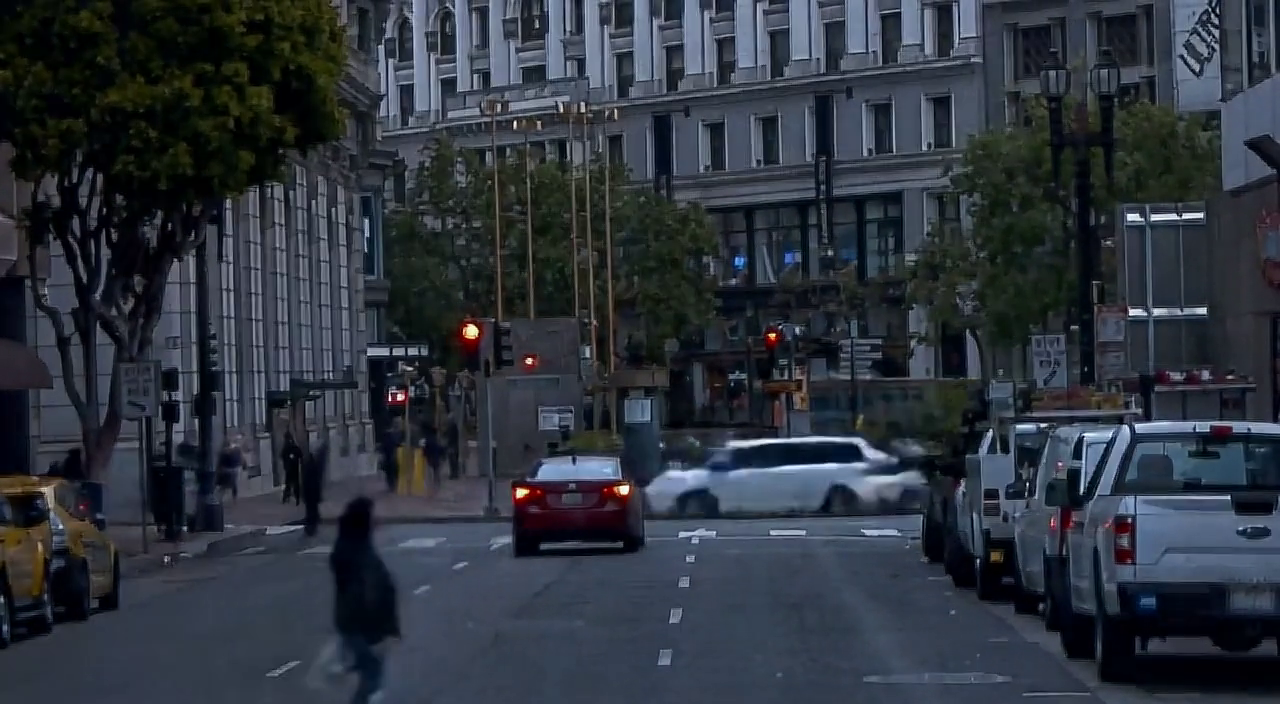}{0.15}{0}{0.75}{0.55}
& \imgbox{0.17\linewidth}{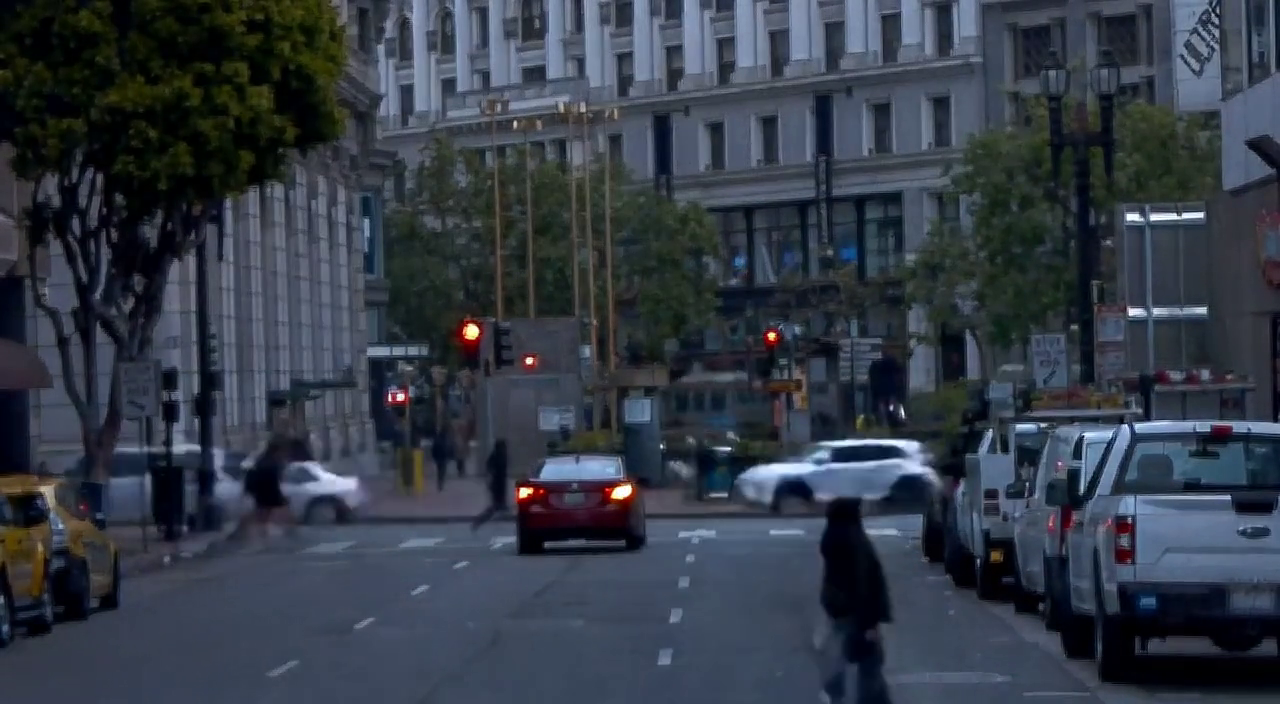}{0.15}{0}{0.75}{0.55}
\\
\end{tabular}
}

\vspace{3pt}
\caption{
Scale Change Query. Top: observed evidence frames from previous iteration. Middle: observation estimates produced by GEN3C at next iteration. Bottom: observation estimates produced by AW4RE at next iteration. AW4RE produces high-quality zoom-in views in which fine-grained details remain consistent with the corresponding far-view observation.
}
\label{fig:zoom_in}
\end{figure*}

\begin{figure*}[t]
\centering
\setlength{\tabcolsep}{4pt}

{\footnotesize
\begin{tabular}{c|ccccc}
 & $o^{(1)}_1$ & $o^{(1)}_{20}$ & $o^{(1)}_{50}$ & $o^{(1)}_{70}$ & $o^{(1)}_{100}$ \\
\rotatebox{90}{Evidence}
&  \imgbox{0.17\linewidth}{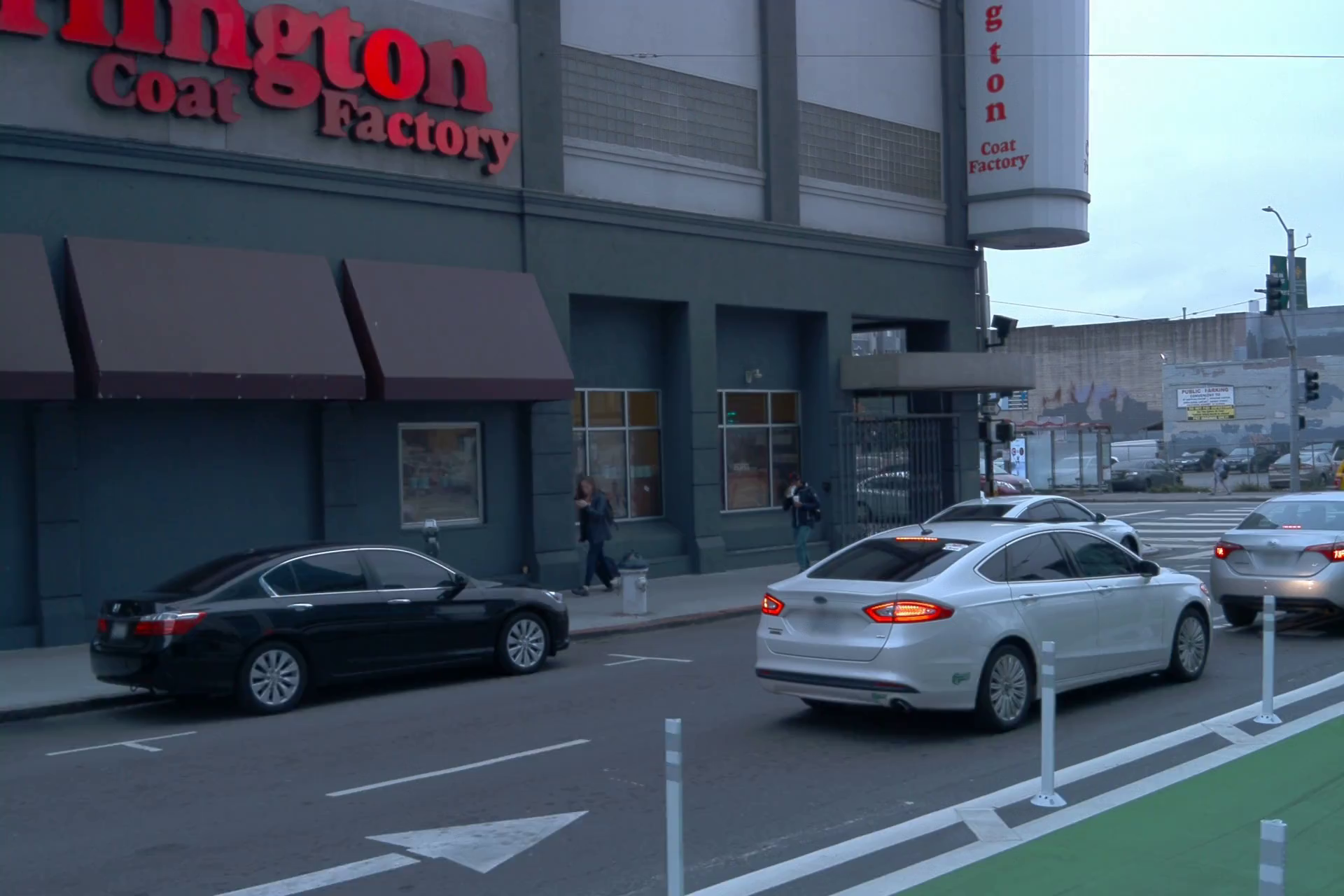}
{0.65}{0.4}{1}{0.7}
&  \imgbox{0.17\linewidth}{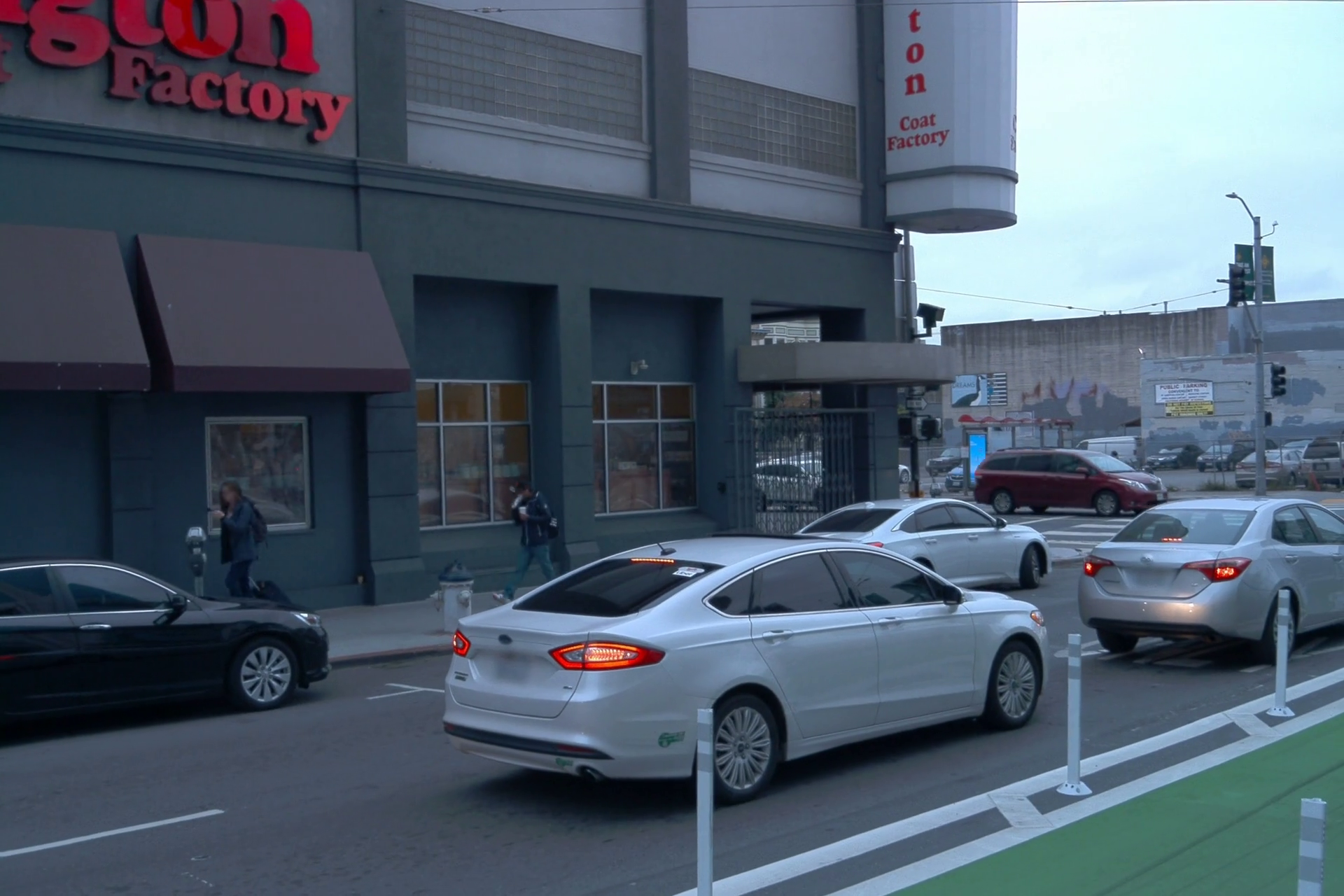}{0.65}{0.4}{1}{0.7}
&  \imgbox{0.17\linewidth}{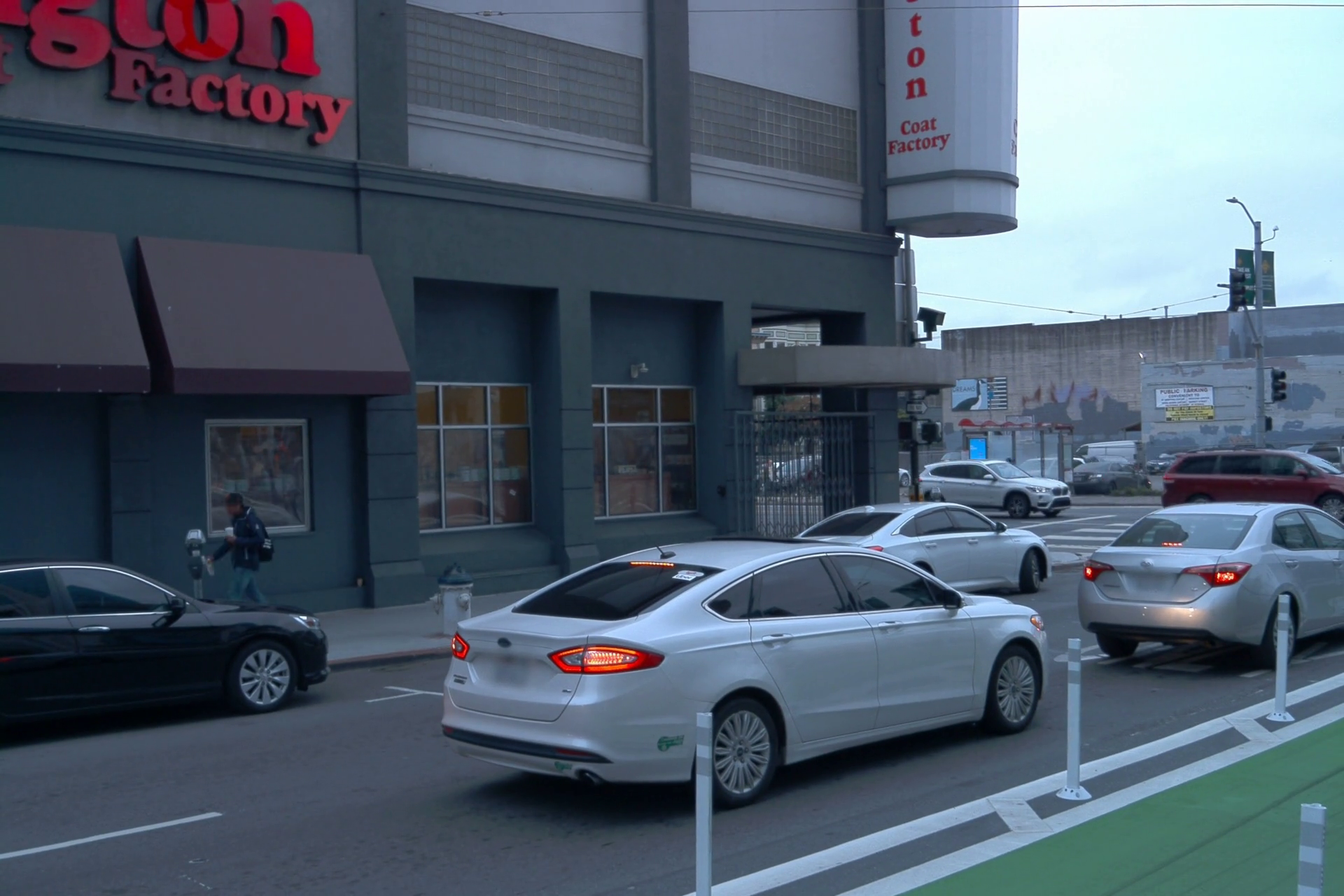}{0.65}{0.4}{1}{0.7}
&  \imgbox{0.17\linewidth}{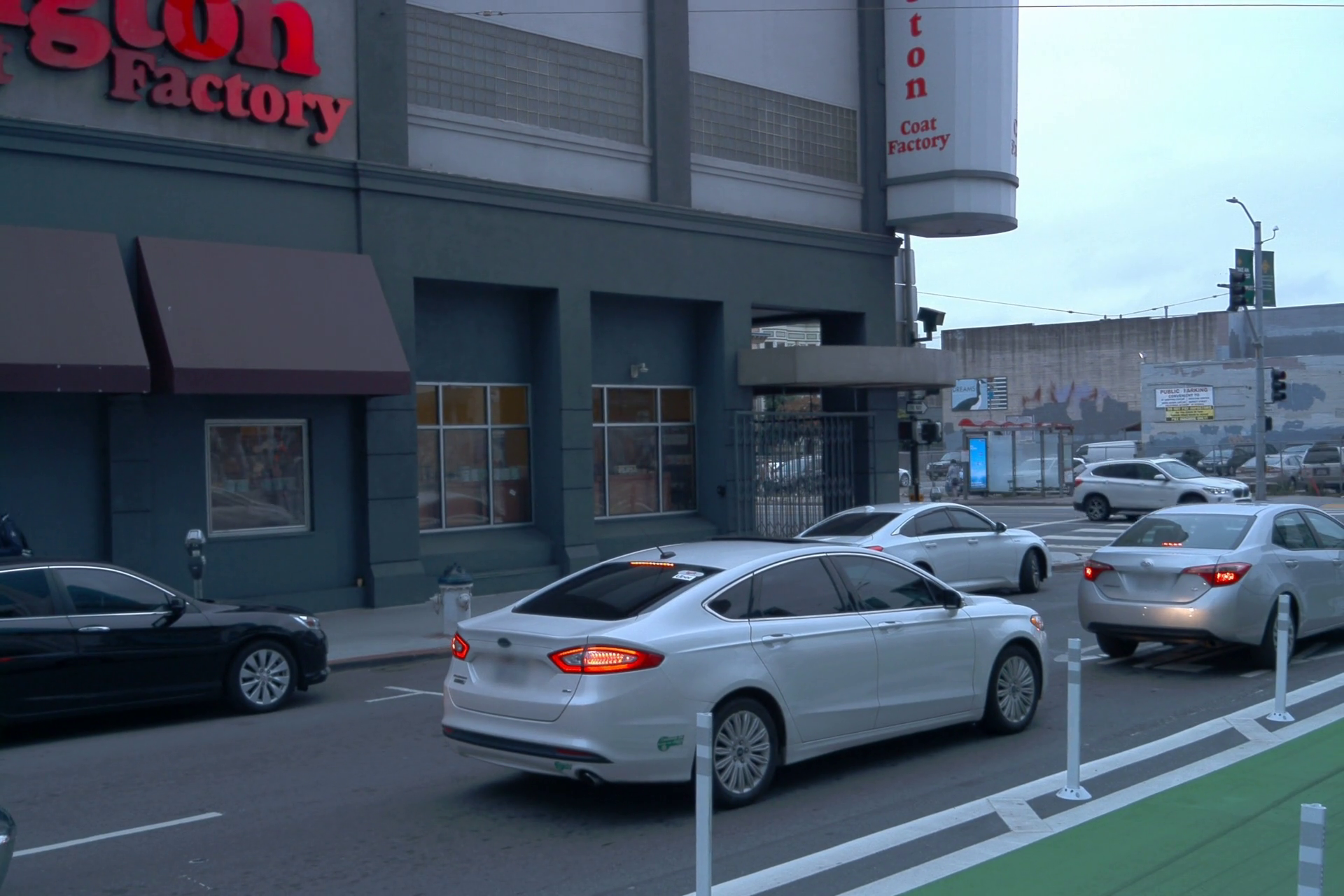}{0.65}{0.4}{1}{0.7}
&  \imgbox{0.17\linewidth}{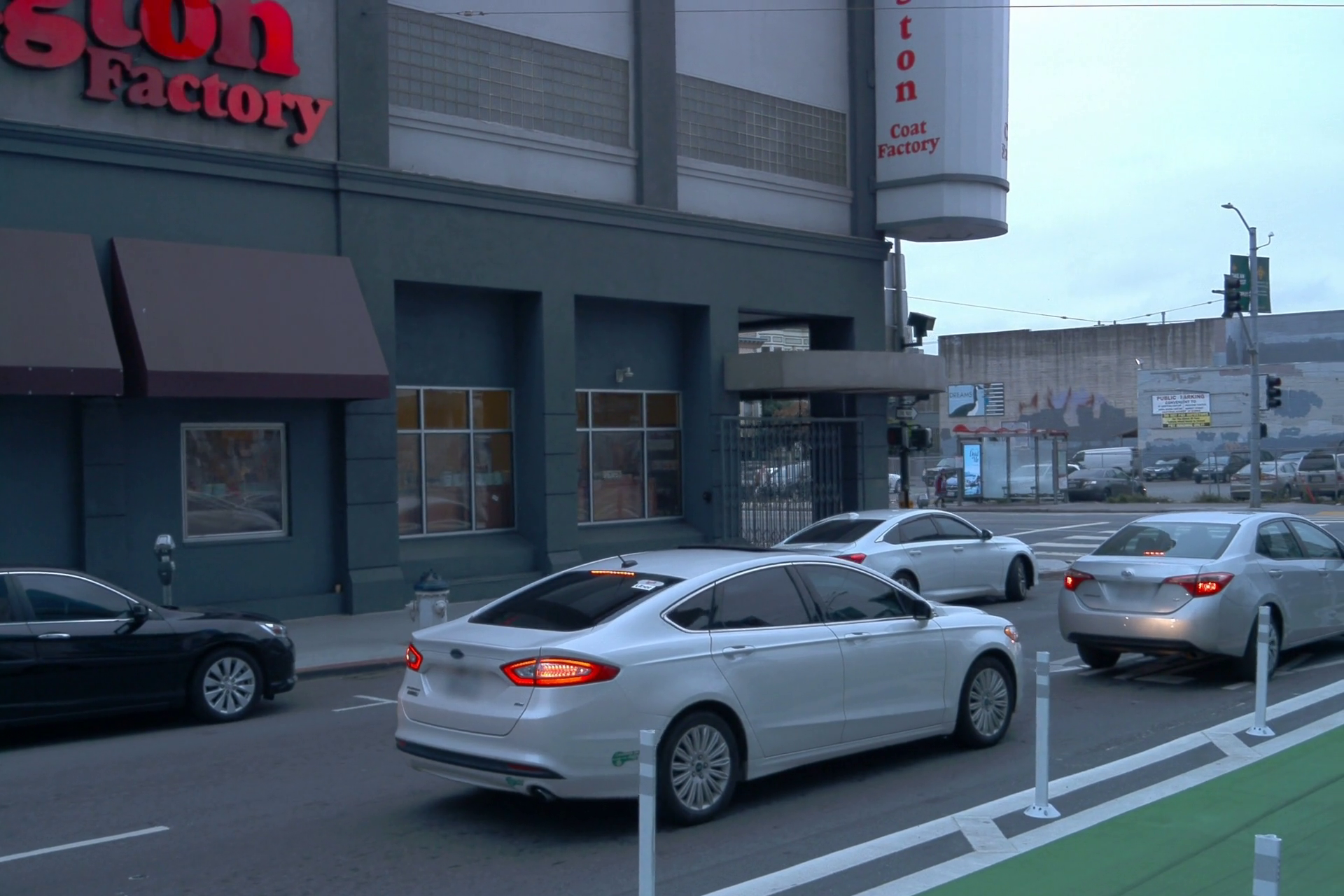}{0.65}{0.4}{1}{0.7}
\\
& \multicolumn{5}{c}{} \\
 & \multicolumn{5}{c}{$\mathcal{D}^{(1)} = \{(\{o^{(1)}_t\}_{t=1}^{121}, \{a^{(1)}_t\}_{t=1}^{121})\}$, $a^{(2)} = \textit{corner-focused view trajectory}$} \\
 & \multicolumn{5}{c}{} \\
 & $o^{(2)}_1$ & $o^{(2)}_{20}$ & $o^{(2)}_{50}$ & $o^{(2)}_{70}$ & $o^{(2)}_{100}$ \\
\rotatebox{90}{GEN3C}
& \imgbox{0.17\linewidth}{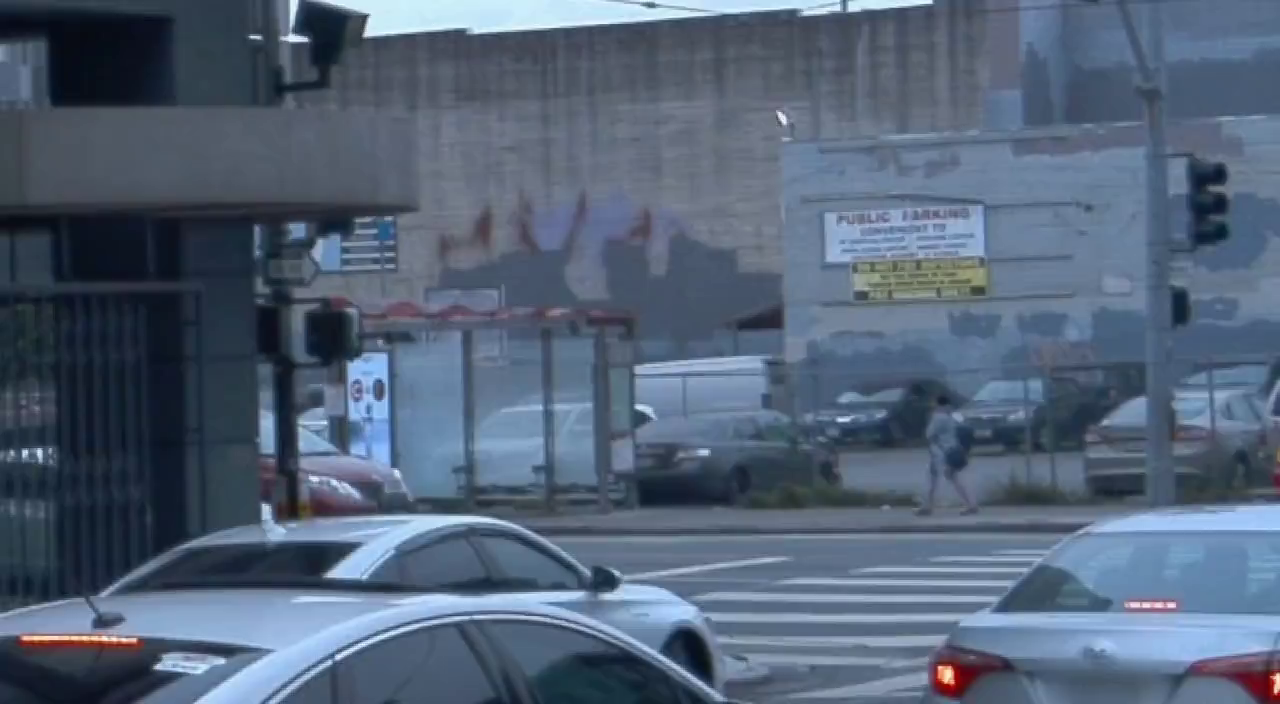}{0.05}{0.05}{1}{0.7}
& \imgbox{0.17\linewidth}{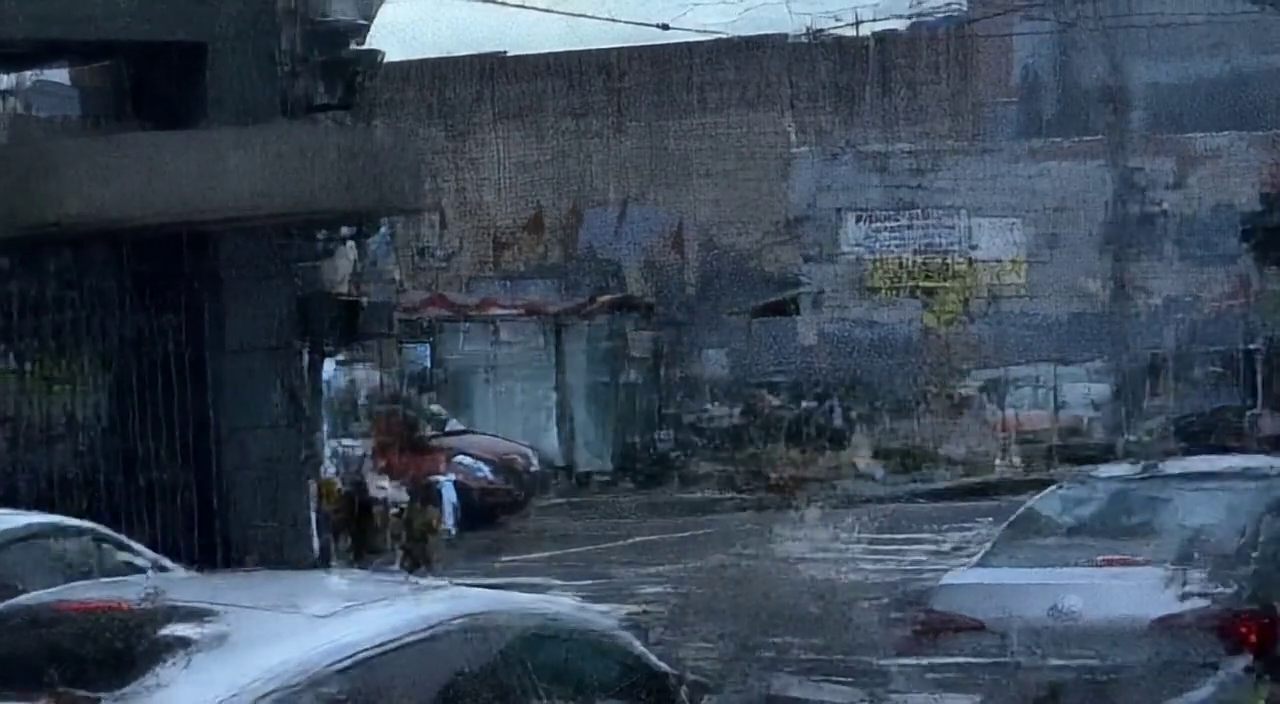}{0.05}{0.05}{1}{0.7}
&\imgbox{0.17\linewidth}{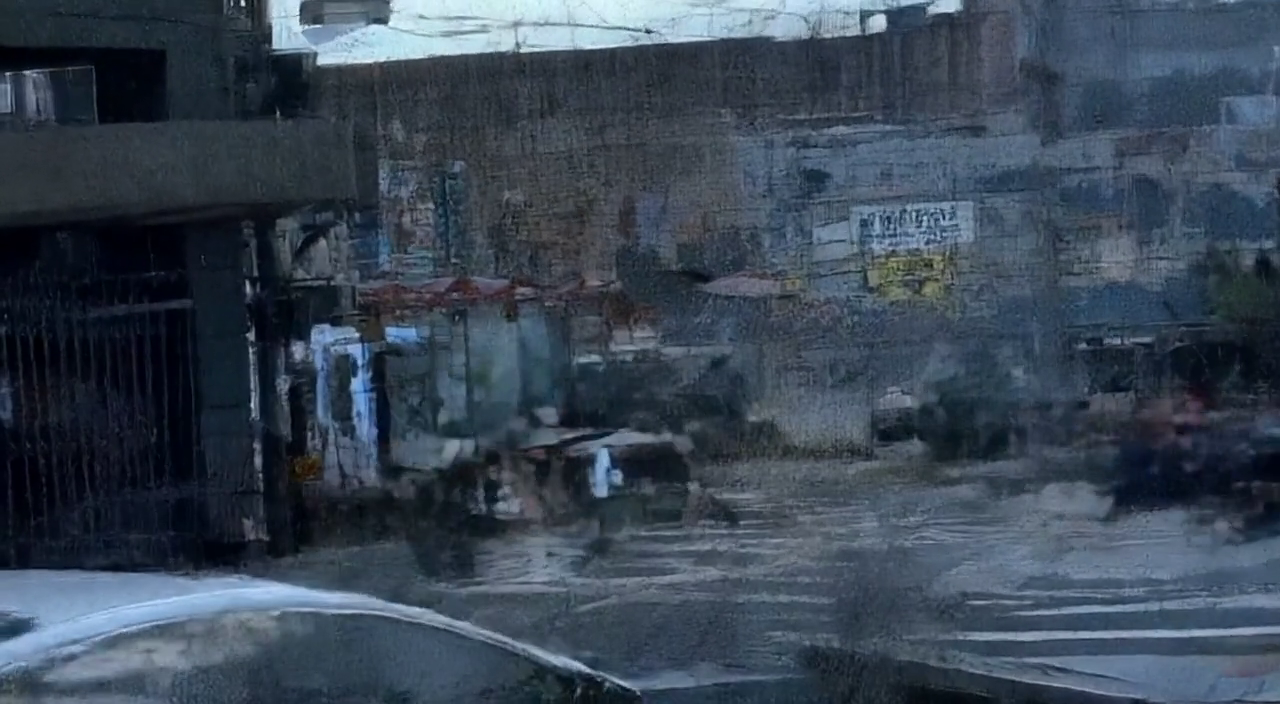}{0.05}{0.05}{1}{0.7}
&\imgbox{0.17\linewidth}{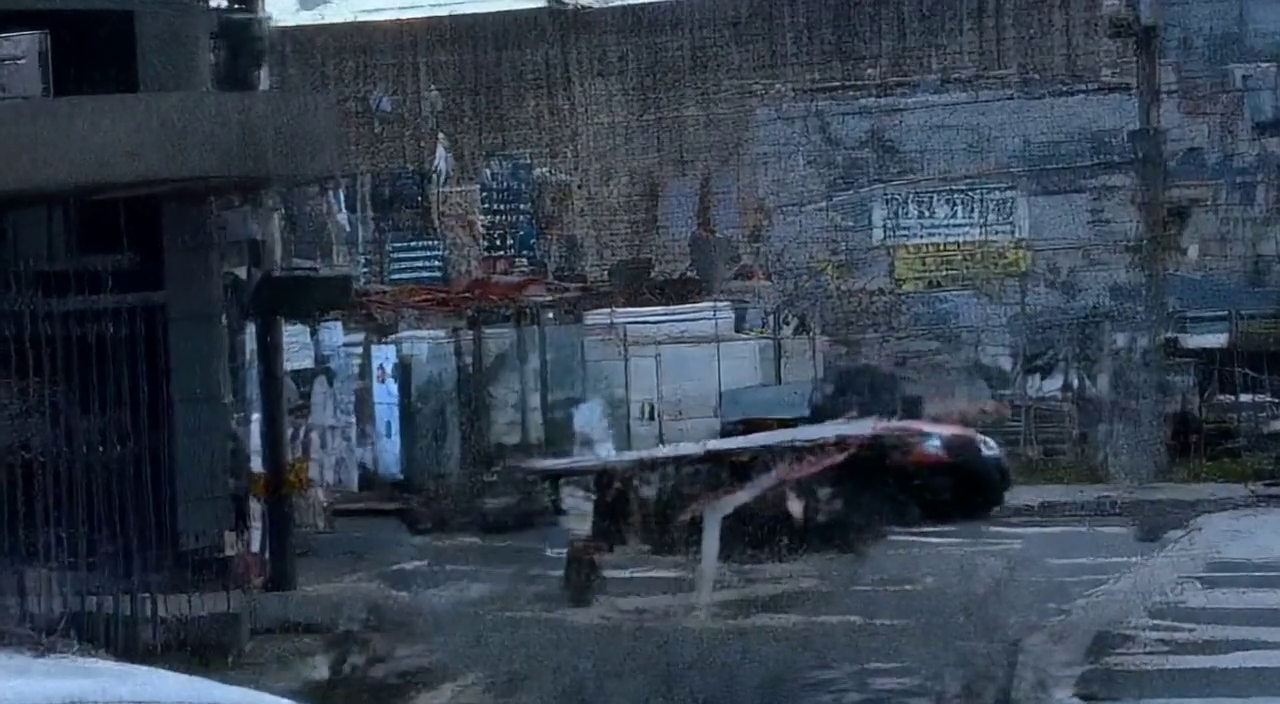}{0.05}{0.05}{1}{0.7}
& \imgbox{0.17\linewidth}{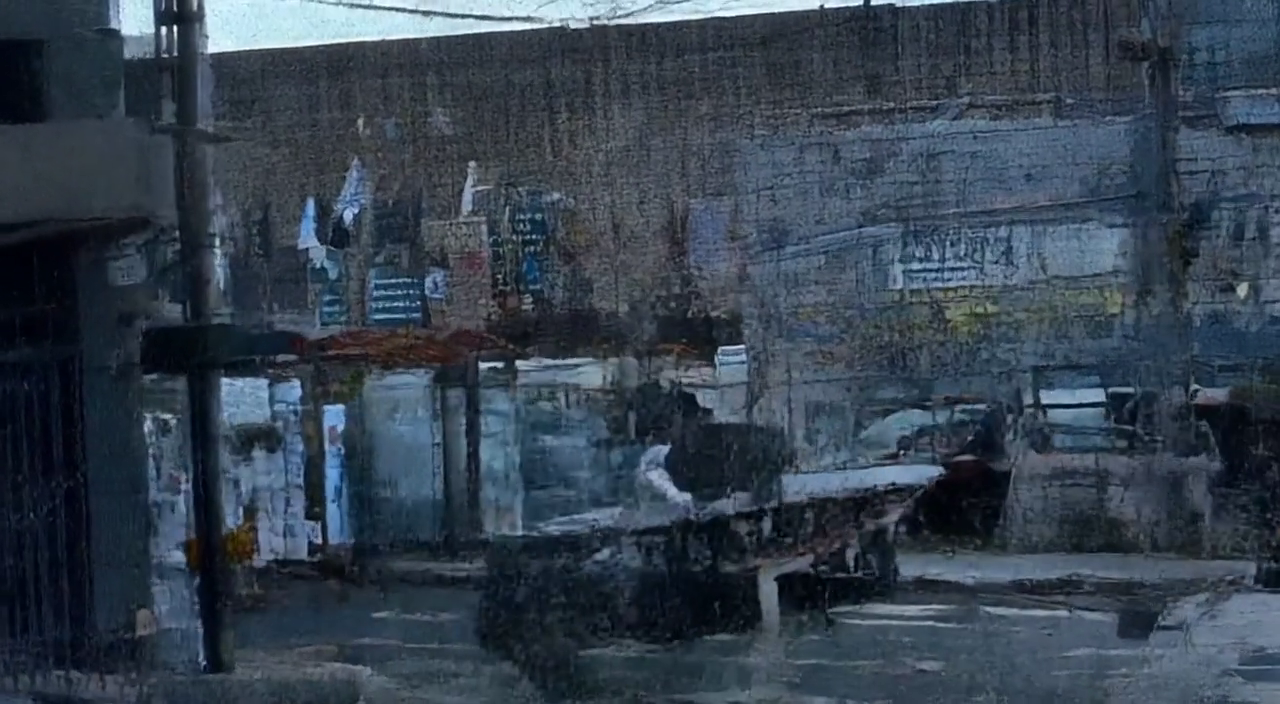}{0.05}{0.05}{1}{0.7}
\\

\rotatebox{90}{AW4RE}
&  \imgbox{0.17\linewidth}{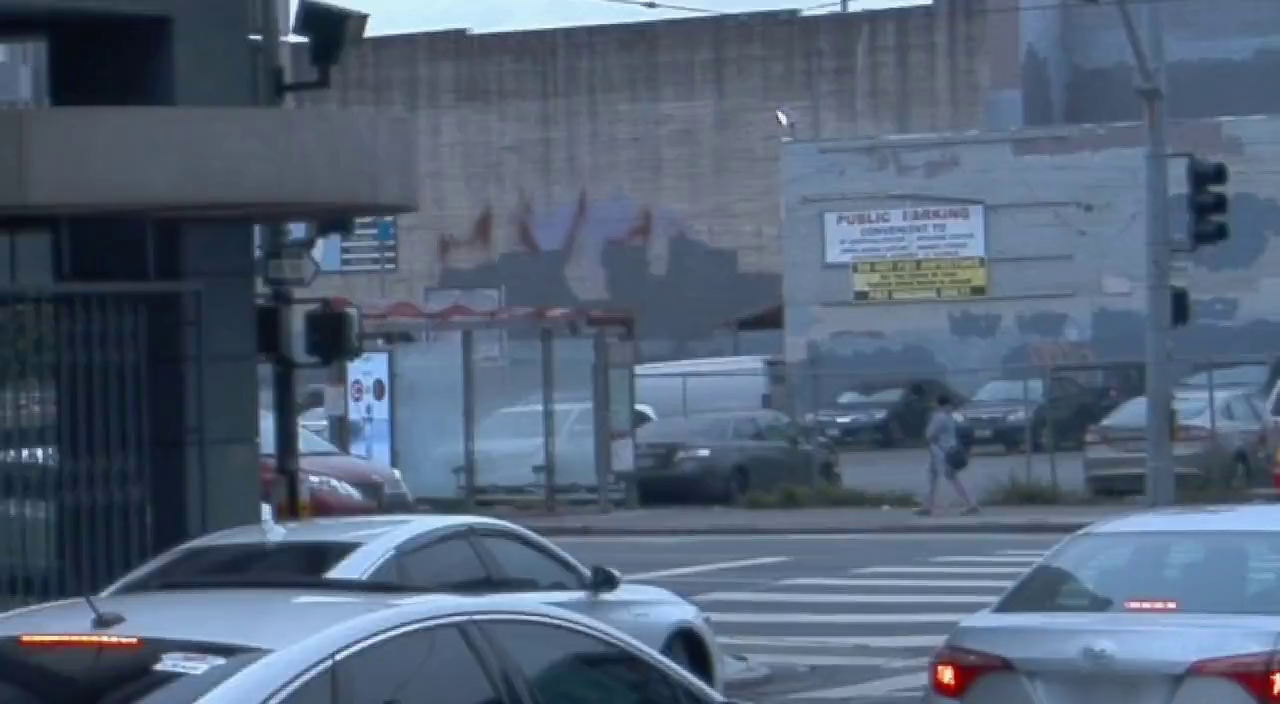}{0.25}{0.2}{1}{0.7}
& \imgbox{0.17\linewidth}{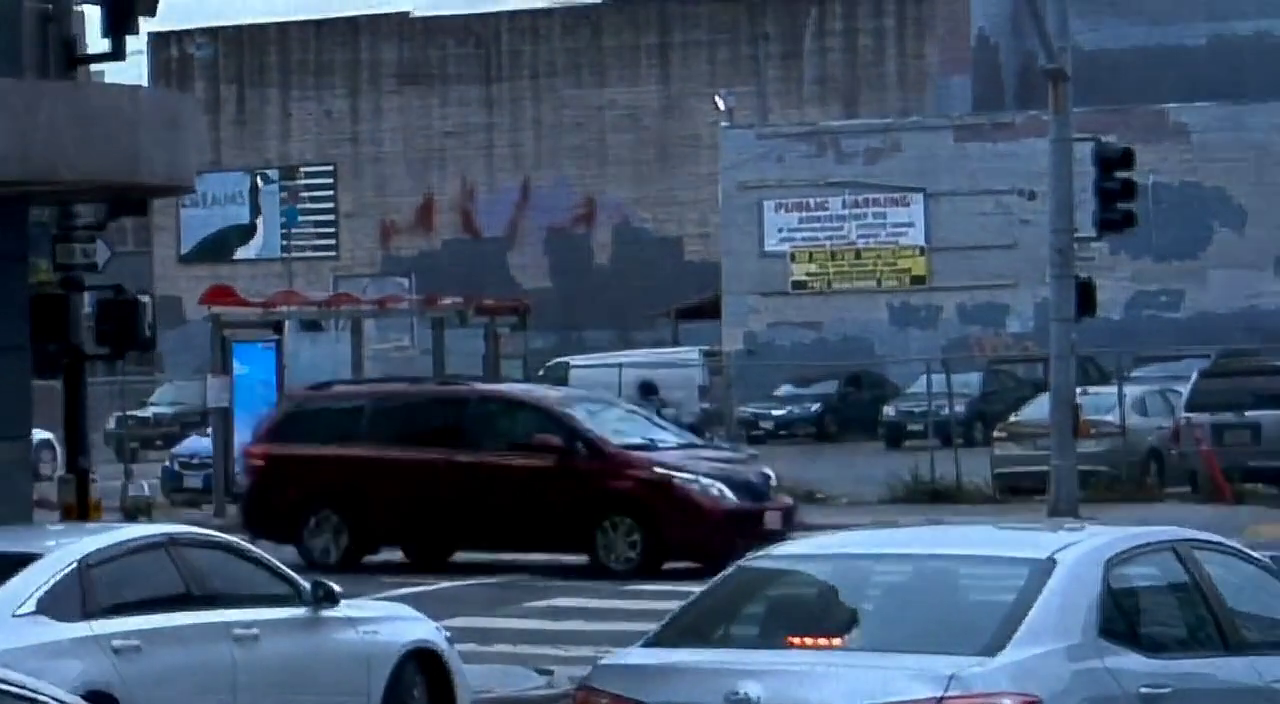}{0.15}{0.05}{0.6}{0.7}
& \imgbox{0.17\linewidth}{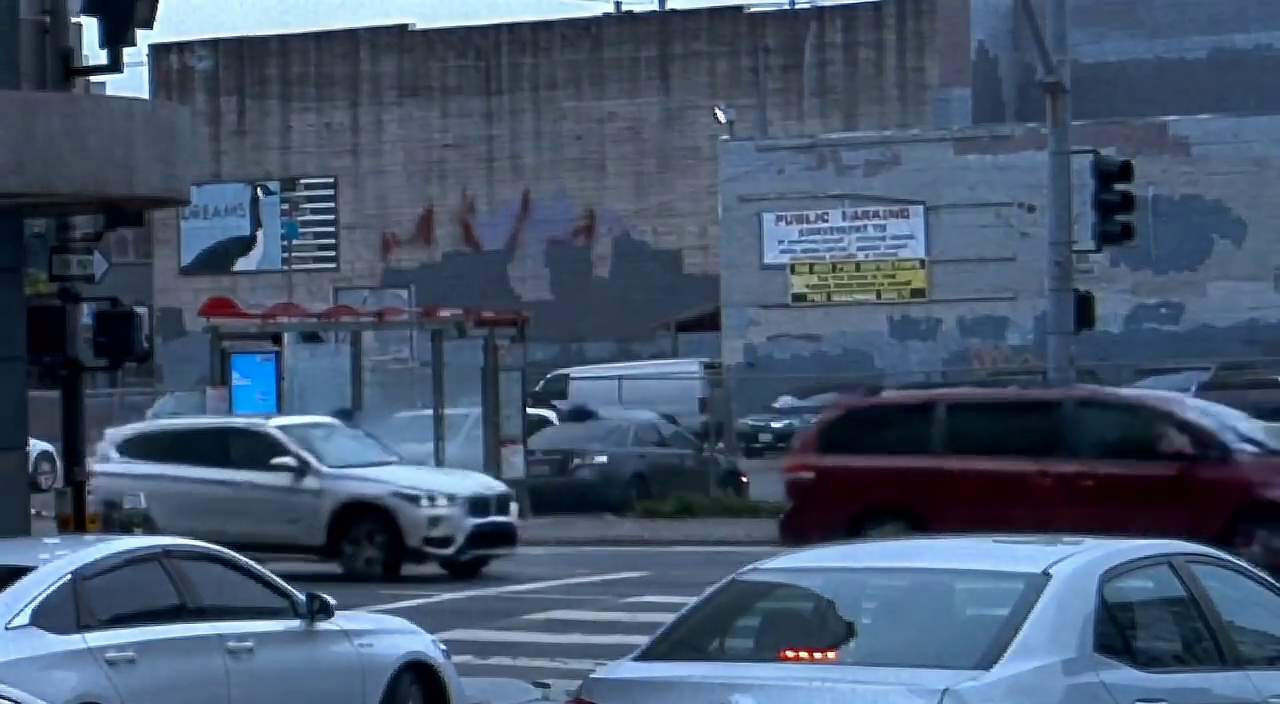}{0.05}{0.05}{1}{0.7}
& \imgbox{0.17\linewidth}{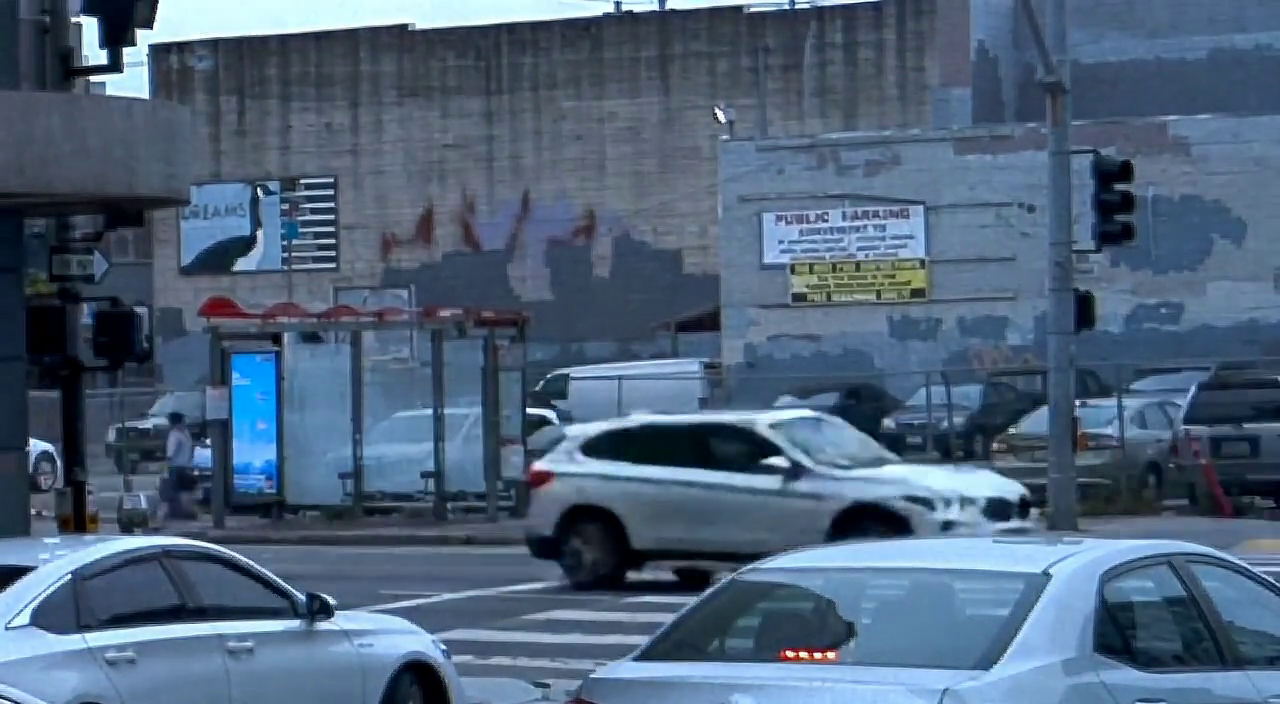}{0.35}{0.15}{0.85}{0.5}
& \imgbox{0.17\linewidth}{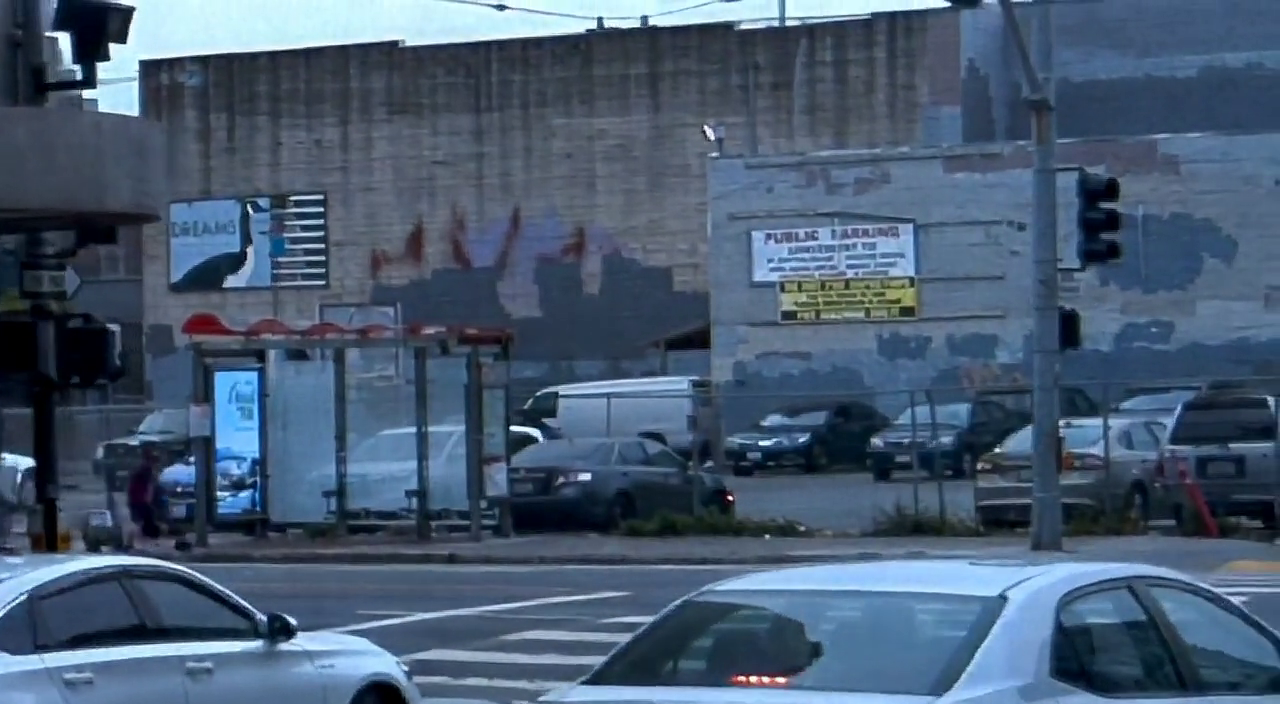}{0.15}{0.2}{1}{0.7}
\\
\end{tabular}
}

\vspace{3pt}
\caption{
Counterfactual View Query. Top: observed evidence frames from previous iteration. Middle: observation estimates produced by GEN3C at next iteration. Bottom: observation estimates produced by AW4RE at next iteration. Despite significant viewpoint overlap, GEN3C struggles, whereas AW4RE generates improved predictions by leveraging 4D-informed evidence retrieval.
}
\label{fig:corner_view}
\end{figure*}

\vspace*{-.13in}
\paragraph{Sparse Geometric Support}
We evaluate a scale change camera action (zoom in) where the query view overlaps with past observations but lacks sufficient information to fully constrain the scene as presented in Fig. \ref{fig:zoom_in}. Although viewpoints overlap, changes in scale and field of view result in sparse and incomplete geometric support, yielding a limited point cloud when evidence is lifted into 3D. GEN3C conditions generation only on overlapping frames at the same timestamp and does not account for the quality or coverage of geometric support; consequently, it hallucinates structure and fine details in under-constrained regions of the zoomed-in view. In contrast, our 4D-informed model explicitly reasons about geometric support when selecting evidence, aggregating relevant observations across time and viewpoints to maximize coverage of the query view. By projecting this structured but sparse evidence into the query camera, our method constrains generation where support exists and limits hallucination elsewhere, preserving spatial layout and scale consistency under scale change actions.

We further evaluate the models on a corner view of a street scene as presented in Fig.~\ref{fig:corner_view}, where the query camera observes the environment from a previously unseen oblique angle. While parts of the scene are visible in past observations, the change in viewpoint results in limited overlap and severe occlusions, producing a highly sparse and fragmented point cloud when evidence is lifted into 3D. In this setting, time-local conditioning provides little guidance: GEN3C relies on coincident frames and consequently hallucinates structure in regions that are weakly constrained by geometry. In contrast, our 4D-informed model retrieves geometrically relevant evidence across time and viewpoints, even when individual observations contribute only partial support. By projecting this aggregated evidence into the corner view, the model identifies which regions are grounded and which remain uncertain, enabling more stable geometry and consistent scene layout despite sparse support.

In these experiments, GT frames are not available and direct comparison is therefore infeasible. Table \ref{tab:sparse_geometric} summarizes evaluation in this setting. In this regime, we focus on evidence-backed regions and temporal consistency. In both cases of camera action, AW4RE achieves substantially higher evidence-region PSNR and lower T-LPIPS than GEN3C. Together, these results indicate that AW4RE remains faithful to available evidence when generating under sparse constraints, preserving supported structure while avoiding spurious hallucination.

\begin{table}[htbp]
\centering
\caption{Sparse Geometric Support Evaluation}
\label{tab:sparse_geometric}
\setlength{\tabcolsep}{8pt}
\resizebox{\textwidth}{!}{%
\begin{tabular}{@{}llccc@{}}
\toprule
 & & \multicolumn{1}{c}{\textbf{Evidence Metric}} & \phantom{abc} & \multicolumn{1}{c}{\textbf{Temporal Consistency}} \\
\cmidrule{3-3} \cmidrule{5-5}
\textbf{Camera Action} & \textbf{Model} & \textbf{PSNR} $\uparrow$ & & \textbf{Full T-LPIPS} $\downarrow$ \\ \midrule
\textbf{Scale Change Query} & AW4RE  & \textbf{21.106} & & \textbf{0.0508} \\
                & GEN3C & 15.847          & & 0.0534          \\ \midrule
\textbf{Counterfactual View Query} & AW4RE  & \textbf{19.504} & & \textbf{0.0991} \\
                & GEN3C & 12.538          & & 0.1453          \\ \bottomrule
\end{tabular}%
}
\end{table}

\vspace*{-.063in}
\paragraph{Observed Properties.}
Our experimental results indicate that AW4RE exhibits the following properties, which are critical for serving as a surrogate environment for awareness-driven reasoning under partial observability.

\vspace*{-.13in}
\paragraph{Property 1 (Multi-view and spatio-temporal consistency).}
Across diverse viewpoint changes and temporal queries, AW4RE produces observations that remain geometrically consistent with a single underlying scene. Predictions generated from different camera positions, orientations, or times preserve stable structure and object relationships, rather than drifting across views or time.

\vspace*{-.13in}
\paragraph{Property 2 (Multi-scale coherence).}
AW4RE maintains coherence across scale. Fine-grained details inferred under zoom-in actions remain compatible with coarse contextual structure observed from distant viewpoints, enabling smooth transitions between global context and local detail without introducing scale-dependent artifacts.

\vspace*{-.13in}
\paragraph{Property 3 (Awareness-driven inference).}
Generated observations are explicitly constrained by prior evidence. AW4RE distinguishes between evidence-supported and unsupported regions, grounding predictions where geometric support exists and limiting hallucination elsewhere, rather than optimizing solely for visual plausibility.

\vspace*{-.13in}
\paragraph{Property 4 (Counterfactual, policy-evaluable behavior).}
AW4RE supports counterfactual sensing by generating observations corresponding to alternative camera actions, including viewpoints, scales, and times that were never directly observed. These counterfactual predictions remain stable and informative, enabling comparative evaluation of sensing decisions.

\vspace*{-.13in}
\paragraph{Property 5 (Environment-agnostic generalization).}
The model generalizes across environments without modification. When conditioned on evidence from a new scene, AW4RE adapts its predictions accordingly, supporting counterfactual reasoning without manual reconstruction, environment-specific tuning, or retraining.

Together, these observed properties distinguish AW4RE from models designed primarily for visual generation, reconstruction, or short-horizon control, and demonstrate its suitability as an awareness-centric world model for exploration and camera-action evaluation.

\vspace*{-.13in}
\section{Discussion and Conclusion}
\vspace*{-.13in}
We reframed physical awareness as a decision-making problem under partial observability and identified the limitations of existing approaches. We introduced the first active world model with 4D-informed retrieval designed to support counterfactual spatial reasoning and awareness-driven inference.
This perspective enables learning sensing and attention policies without unsafe real-world exploration and provides a foundation for scalable spatial reasoning in complex environments. As future work, we aim to jointly train a model that integrates both the 4D-informed State Estimator module and the Evidence-backed Observation module in an end-to-end manner.

\newpage
\bibliographystyle{iclr2026_conference}
\bibliography{references}

\end{document}